\documentclass[preprint,12pt]{elsarticle}

\usepackage{lineno,hyperref}
\hypersetup{pdftitle={FairPGT: Partitioned Graph Transformers for Fair Graph Clustering},
            pdfauthor={Siamak Ghodsi et al.}}
\usepackage[utf8]{inputenc} %
\usepackage[T1]{fontenc}    %
\usepackage{lmodern}
\usepackage{url}            %
\usepackage{booktabs}       %
\usepackage{amsfonts}       %
\usepackage{nicefrac}       %
\usepackage{microtype}      %
\usepackage{longtable}      %
\usepackage{graphicx}
\usepackage{amsmath, amssymb, amsthm}
\usepackage{algorithm}
\usepackage{algpseudocode} %
\usepackage[table]{xcolor}
\definecolor{boston}{rgb}{0.8, 0.0, 0.0}
\usepackage{tabularx,array,adjustbox,pifont,amsthm,etoolbox,enumitem,comment}
\definecolor{exprow}{RGB}{237,243,252}
\definecolor{fairbg}{RGB}{245,240,250}
\newcommand{\cmark}{\ding{51}}
\newcommand{\xmark}{\ding{55}}

\newcommand{\std}[1]{{\scriptsize${\pm\text{#1}}$}}
\usepackage{multirow}
\usepackage{amsfonts}
\usepackage{mathrsfs}  
\usepackage{bm}
\usepackage{setspace}
\usepackage{subcaption}
\usepackage{mathtools}
\usepackage[margin=2.5cm]{geometry}
\usepackage{tikz}
\usepackage{makecell}

\newtheorem{theorem}{Theorem}
\newtheorem{proposition}{Proposition}
\newtheorem{remark}{Remark}
\newtheorem{definition}{Definition}

\newcommand*\linenomathpatch[1]{%
  \cspreto{#1}{\linenomath}%
  \cspreto{#1*}{\linenomath}%
  \csappto{end#1}{\endlinenomath}%
  \csappto{end#1*}{\endlinenomath}%
}
\linenomathpatch{equation}
\linenomathpatch{gather}
\linenomathpatch{multline}
\linenomathpatch{align}
\linenomathpatch{alignat}
\linenomathpatch{flalign}

\makeatletter
\newcommand{\heading}[1]%
{\par\vskip 1.5ex \@plus .2ex
  \hangindent=1em
  \noindent\makebox[1em][l]{$\,\bullet$}\textit{#1}%
  \par\vskip 1.5ex \@plus .2ex
  \@afterheading}
\makeatother

\definecolor{amc}{RGB}{0,102,204}

\date{}

\begin{document}
	\begin{frontmatter}
		
    \title{\textbf{FairPGT: Partitioned Graph Transformers\\for Dual-Level Fair Graph Clustering}}
		
    \author[1,2,3]{Siamak Ghodsi\corref{corauthor}}
    \ead{ghodsi@l3s.de}

    \author[4]{Amjad Seyedi}
    \ead{seyedamjad.seyedi@umons.ac.be}

    \author[5]{Sahar Tahmasebi}
    \ead{sahar.tahmasebi@tib.eu}

    \author[1,6]{Eirini Ntoutsi}
    \ead{eirini.ntoutsi@unibw.de}

    \author[1,2,5]{Sören Auer}
    \ead{auer@tib.eu}

    \address[1]{L3S Research Center, Hannover, Germany}
    \address[2]{Leibniz University Hannover, Hannover, Germany}
    \address[3]{Department of Mathematics and Informatics, Free University Berlin, Berlin, Germany}
    \address[4]{Department of Mathematics and Operational Research, University of Mons, Mons, Belgium}
    \address[5]{TIB -- Leibniz Information Centre for Science and Technology, Hannover, Germany}
    \address[6]{Research Institute CODE, University of the Bundeswehr Munich, Munich, Germany}
    \cortext[corauthor]{Corresponding author}
		
\begin{abstract}

Clustering on attributed graphs can produce demographically biased clusters, particularly when communities align with protected groups. Graph transformers can amplify this bias through same-group attention, while global attention limits scalability. We propose FairPGT, a partitioned graph transformer for fair graph clustering. FairPGT restricts attention to METIS partitions and connects them through routing tokens, which divides the quadratic cost of attention by the number of partitions. A graph-aware self-training target guides clustering, with a provably bounded deviation from the standard DEC target. Fairness is imposed at two levels: a mutual-information term promotes demographic parity in cluster assignments, while a normalized cross-group attention ratio (nCGAR) guides attention toward group neutrality through self-distillation. We prove that the nCGAR correction moves attention toward neutrality and establish a condition preventing overshooting. Across seven benchmarks, FairPGT achieves the best clustering performance and statistical parity on the three largest graphs (Credit, Pokec-n, Pokec-z), with a statistical-parity gap approximately ten times lower than DMoN, and is Pareto-dominated on only one dataset. On controlled stochastic block models, FairPGT attains 97\% of the derived upper bound on NMI for exactly fair clustering when communities and protected groups coincide. These graphs further show that, under strong alignment, assignment-level fairness alone makes the attention group-heterophilous, whereas attention-level fairness alone improves neutrality without ensuring fair clusters. Code for reproducibility is available at \href{{https://github.com/SiamakGhodsi/FairPGT}}{https://github.com/SiamakGhodsi/FairPGT}.

\end{abstract}

\begin{keyword}
    Fair Graph Clustering, Graph Transformers, Algorithmic Fairness, Partitioned Attention, Unsupervised Representation Learning
\end{keyword}
\end{frontmatter}

\section{Introduction}
When a bank segments its customers or a platform groups its users, the resulting clusters decide who is offered what. Graph clustering, closely related to community detection, groups the nodes of a graph into structurally coherent communities and is a fundamental unsupervised task in network analysis, with applications in fraud analysis, bioinformatics, and recommendation~\citep{fortunato2016community,newman2006modularity,DBLP:conf/ijcai/JuY0L00024,DBLP:journals/tkde/DongMWCL23,DBLP:journals/tnn/LiuYZLWLTL24}. When nodes carry protected attributes such as gender, age, or nationality, the communities a clustering finds can line up with demographic groups, and because clusters are used downstream, to target services or to segment users, such alignment can turn into biased decisions~\citep{DBLP:conf/nips/BeraCFN19,DBLP:conf/pakdd/GhodsiSN24}. The problem is hardest in unsupervised graph learning, where no label tells useful structure apart from structure that merely reflects the protected attribute.

Bias can enter a clustering pipeline at two points. First, at the representation level: node representations may encode group membership even when the protected attribute is removed from the input, because node features and graph structure act as proxies for it~\citep{DBLP:conf/innovations/DworkHPRZ12,DBLP:journals/tkdd/ChenRPTWYKDA24}. In message-passing graph neural networks, aggregation over neighborhoods dominated by one group has been shown to amplify this signal~\citep{DBLP:journals/corr/fmp,DBLP:conf/log/LinKCT23}. Similarly, graph transformers aggregate information through attention~\citep{kreuzer2021rethinking,DBLP:conf/nips/RampasekGDLWB22,DBLP:conf/icml/ChenOB22}, and they are exposed to the same effect in a more flexible form: if attention concentrates on nodes of the same group, each representation becomes a summary of its own group. Moreover, at the assignment level, clusters can be demographically imbalanced even when the representations are reasonable; this is the level addressed by most classical fair clustering methods~\citep{chierichetti2017fair,DBLP:conf/nips/BeraCFN19,kleindessner2019fair}.

\begin{figure}[t]
\centering
\includegraphics[width=\linewidth]{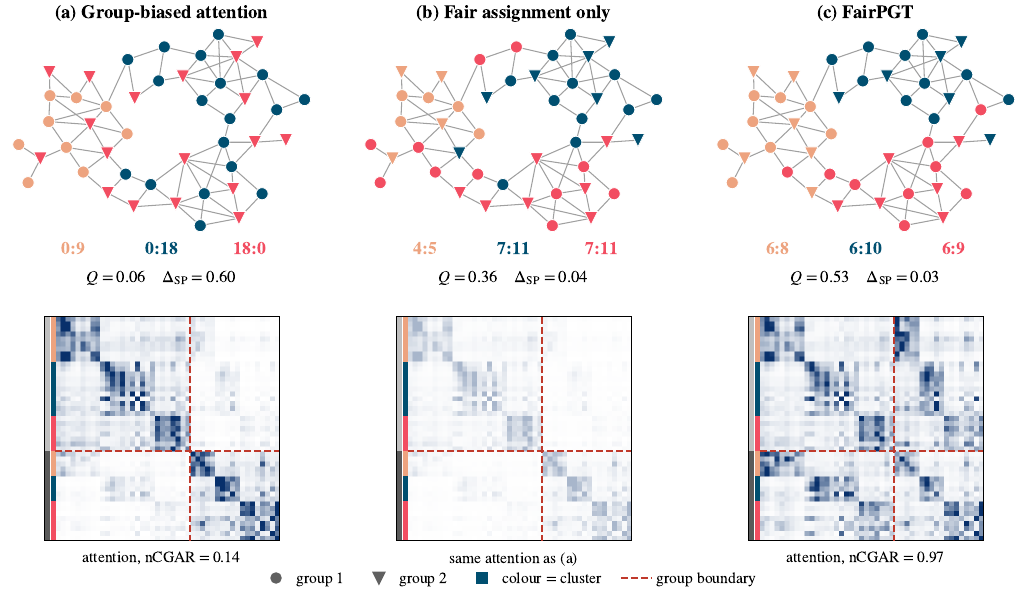}
\caption{\textbf{Why fairness is needed in both attention and assignments.}
Toy graph with 45 nodes, three communities, and two protected groups mixed within each community. Colours denote clusters, shapes denote groups, and numbers give the group counts (triangles:circles) of each cluster. Spectral coordinates and a noisy group-proxy feature feed one attention layer, whose representations are clustered by $k$-means ($K=3$).
(a)~Same-group attention ($\mathrm{nCGAR}=0.14$) splits the clusters by group.
(b)~Enforcing the population group shares on the same representations balances the clusters but fragments the communities.
(c)~Replacing the attention by the fixed point of the CGAR target (Eq.~\eqref{eq:cgar_correction}; $\mathrm{nCGAR}=0.97$), the attention toward which FairPGT's self-distillation drives its encoder, and applying the same group shares yields clusters that are both balanced and cohesive.
Panels (b) and (c) isolate the two levels by direct intervention, so that their effects can be seen without training; the outcome is the same for attention temperatures 0.3, 0.5, and 1.0.}
\label{fig:motivation}
\end{figure}

\autoref{fig:motivation} illustrates both these levels concretely on a small graph with 45 nodes and depicts why balancing cluster assignments alone may be insufficient. In this toy graph, the protected groups are mixed within the communities, but group-correlated features draw attention toward same-group nodes and pull clusters away from the communities. Enforcing group proportions on the assignments balances the clusters while disrupting their community structure. Rebalancing attention alongside assignment-level fairness yields clusters that are both balanced and structurally cohesive.

Two observations guide our design. First, attention and assignments are separate entry points for group information, and fixing one does not explicitly fix the other: with a fixed encoder, fair assignments have to work against biased representations, as in panel~(b), and we show in \autoref{sec:sbm} that an encoder trained with assignment-level fairness alone learns attention biased toward the other group, while neutral attention alone leaves the clusters unfair. Second, same-group attention has to be judged relative to the group composition of the attended nodes, because even group-agnostic attention gives more total mass to the larger group.

Existing methods cover parts of this picture. On the clustering side, fair clustering methods impose proportionality or individual-fairness constraints in spectral formulations~\citep{kleindessner2019fair,DBLP:conf/nips/GuptaD22,wang2023scalable,DBLP:journals/pr/FengZQPY26}, or they add parity, adversarial, and contrastive regularizations to deep clustering models to remove protected information from learned representations~\citep{DBLP:conf/cvpr/LiZL20,DBLP:conf/aaai/ZhangWRWYMMLJ26,DBLP:journals/pr/JiangZHLHC26}. In addition, our previous factorization-based models made the fairness--utility trade-off explicit at the assignment level~\citep{DBLP:conf/pakdd/GhodsiSN24,DBLP:conf/bigdataconf/GhodsiSQ0N25}. However, all of these methods act on the assignments or on the learned representations, and none of their encoders uses attention, so the group dependence of attention is neither measured nor controlled. In parallel, fair graph learning has developed adversarial, counterfactual, and graph-editing methods to debias graph neural network representations~\citep{DBLP:conf/wsdm/DaiW21,DBLP:conf/uai/AgarwalLZ21,DBLP:conf/www/DongLJL22,DBLP:journals/pr/ChenWCLLZ26,DBLP:journals/nn/WangLGBCCL26,DBLP:journals/ijon/ZhaoHLYPFP25}, but mostly for supervised node classification. Closer to our setting, graph transformers have been used for attributed graph clustering, though without fairness considerations~\citep{DBLP:journals/eswa/ZhangZSRJ25}, while the two recent fair graph transformers of Luo et al., FairGT~\citep{DBLP:conf/ijcai/LuoHY0024} and FairGP~\citep{DBLP:conf/aaai/LuoH0X0025}, still target supervised node classification: FairGT builds fairness-aware structural and multi-hop encodings, and FairGP restricts attention to graph partitions and removes attention across them. Consequently, no existing method measures and controls the group dependence of attention in unsupervised graph clustering. Finally, scalability is a further obstacle: full attention costs $\mathcal{O}(n^2)$ per layer, which is prohibitive on the larger graphs of fair graph learning benchmarks.

To address these gaps, we propose \textbf{FairPGT} (\textbf{Fair} \textbf{P}artitioned \textbf{G}raph \textbf{T}ransformer), a graph transformer for fair unsupervised graph clustering that acts on both levels. Its central idea is that fair clusters and and balanced attention are two separate goals, and that a fair clustering method should reach both: clusters with the group shares of the population, produced by an encoder whose attention is balanced across groups. For scalability, attention is computed within METIS partitions, and routing tokens exchange information across partitions, which reduces the cost of attention by a factor of $N_p$, the number of partitions. At the encoder level, we measure the group dependence of attention with a Cross-Group Attention Ratio (CGAR), normalized by the group composition of each partition so that a value of $1$ corresponds to group-agnostic attention. The encoder is trained to keep this ratio close to $1$ through self-distillation: at every training step, its current attention is rebalanced across groups and used as a target for the attention itself. The protected attribute therefore acts only as a training-time regularizer, and the transformer never takes it as input. At the assignment level, a DEC-style clustering head~\citep{xie2016dec} is trained toward a graph-aware target with modularity-based structural guidance, together with a group-fairness term and an information term that keeps the assignments informative. Our main contributions can be listed as follows:
\begin{itemize}
  \item We propose FairPGT, a partitioned graph transformer for unsupervised clustering that restricts attention to METIS partitions, connects them with routing tokens, and clusters with a graph-aware self-training target. Partitioning divides the quadratic cost of attention by the number of partitions and, in practice, reduces peak memory by an order of magnitude (\autoref{sec:method}).

  \item We introduce two complementary fairness terms: an assignment-level term that makes the clusters independent of the protected attribute, and an attention-level self-distillation term built on nCGAR, a normalized cross-group attention ratio that also serves as a diagnostic of the trained encoder. We prove that the nCGAR correction moves the ratio toward neutrality and give the condition under which it does not overshoot, and we bound how far the graph-aware target can depart from the DEC target (\autoref{sec:theory}).

  \item We conduct a controlled study on stochastic block models in which the alignment between communities and groups is varied, with an information-theoretic bound on the NMI of any exactly fair clustering at full alignment. FairPGT stays fair at every alignment and reaches 97\% of the bound (\autoref{sec:sbm}).

  \item We benchmark 14 methods on seven graphs of up to 67k nodes, with binary and four-group protected attributes, under one label-free protocol (fixed hyperparameters, last-epoch selection, collapse detection). FairPGT achieves the best NMI and the best statistical parity on the three largest graphs. 
\end{itemize}

The remainder of the paper is organized as follows. \autoref{sec:related} reviews related work. \autoref{sec:method} presents the proposed FairPGT model and its theoretical properties. \autoref{sec:experiments} reports the experimental results and discussions, and \autoref{sec:conclusion} concludes the paper.

\section{Related Work}
\label{sec:related}

To position our work within the existing literature, we review three main research directions, as summarized in \autoref{tab:comparison}: deep graph clustering and graph transformers, fair clustering, and fairness in deep graph learning.

\begin{table*}[!t]
  \centering
  \caption{Related methods. Task: Clust.\ = node clustering ($^{*}$: clustering of points without a graph), Class.\ = node classification. Unsup.: trained without class labels. Fairness acts on the cluster assignments (Assign.), on learned representations (Repr.), or on attention (Attn.); indiv.\ = individual fairness. Groups: number of protected groups the method is defined for (--: groups not used by the method). Encoder: AE = autoencoder, Contr.\ = contrastive, Part.\ attn = attention within graph partitions, Hop-token attn = attention over a node's multi-hop tokens. FairPGT uses the protected attribute only in its training objective, never as a model input.}
  \label{tab:comparison}
  \scriptsize
  \setlength{\tabcolsep}{2.5pt}
  \renewcommand{\arraystretch}{1.1}
  \adjustbox{max width=\textwidth}{%
  \begin{tabular}{@{} l l c c c c c c l >{\raggedright\arraybackslash}p{3.6cm} @{}}
    \toprule
    & & & & \multicolumn{3}{c}{\textbf{Fairness}} & & & \\
    \cmidrule(lr){5-7}
    \textbf{Method} & \textbf{Venue} & \textbf{Task} & \textbf{Unsup.} & \textbf{Assign.} & \textbf{Repr.} & \textbf{Attn.} & \textbf{Groups} & \textbf{Encoder} & \textbf{Key idea} \\
    \midrule
    \multicolumn{10}{@{}l}{\textit{Deep graph clustering (fairness-agnostic)}} \\
    DMoN~\cite{DBLP:journals/jmlr/TsitsulinPPM23} & JMLR'23 & Clust. & \cmark & & & & -- & GNN & Differentiable modularity \\
    SDCN~\cite{DBLP:conf/www/Bo0SZL020} & WWW'20 & Clust. & \cmark & & & & -- & AE+GCN & AE--GCN fusion, DEC self-training \\
    CCGC~\cite{DBLP:conf/aaai/YangLZWTZLFZ23} & AAAI'23 & Clust. & \cmark & & & & -- & Contr.\ MLP & Cluster-guided positives and negatives \\
    SCGC~\cite{DBLP:journals/tnn/LiuYZLWLTL24} & TNNLS'24 & Clust. & \cmark & & & & -- & Contr.\ MLP & Low-pass filter, cross-view consistency \\
    DinkNet~\cite{DBLP:conf/icml/dinknet23} & ICML'23 & Clust. & \cmark & & & & -- & GNN & Cluster dilation and shrink \\
    DiffPool~\cite{DBLP:conf/nips/DiffPool18} & NeurIPS'18 & Pooling & \xmark & & & & -- & GNN & Hierarchical differentiable pooling \\
    MinCutPool~\cite{DBLP:conf/icml/BianchiGA20} & ICML'20 & Pooling & \cmark & & & & -- & GNN & Relaxed normalized min-cut \\
    \midrule
    \multicolumn{10}{@{}l}{\textit{Fair clustering: classical, spectral, and factorization}} \\
    Fairlets~\cite{chierichetti2017fair} & NeurIPS'17 & Clust.$^{*}$ & \cmark & \cmark & & & 2 & -- & Fair micro-clusters, $k$-center/median \\
    Fair SC~\cite{kleindessner2019fair} & ICML'19 & Clust. & \cmark & \cmark & & & $\ge2$ & Spectral & Proportionality constraint \\
    sFairSC~\cite{wang2023scalable} & AISTATS'23 & Clust. & \cmark & \cmark & & & $\ge2$ & Spectral & Scalable fair spectral clustering \\
    iFairSC~\cite{DBLP:conf/nips/GuptaD22} & NeurIPS'22 & Clust. & \cmark & indiv. & & & -- & Spectral & Representation-graph constraint \\
    iFairNMTF~\cite{DBLP:conf/pakdd/GhodsiSN24} & PAKDD'24 & Clust. & \cmark & indiv. & & & $\ge2$ & NMF & Contrastive individual-fair NMTF \\
    DFNMF~\cite{DBLP:conf/bigdataconf/GhodsiSQ0N25} & BigData'25 & Clust. & \cmark & \cmark & & & $\ge2$ & Deep NMF & Statistical-parity regularized deep NMF \\
    \midrule
    \multicolumn{10}{@{}l}{\textit{Fair deep graph clustering and fair graph transformers}} \\
    Mod.-fair DMoN~\cite{DBLP:conf/icdm/GkartziosPT25} & ICDM'25 & Clust. & \cmark & \cmark & & & 2 & GNN & Group-modularity terms in DMoN \\
    FairGC~\cite{DBLP:conf/aaai/ZhangWRWYMMLJ26} & AAAI'26 & Clust. & \cmark & indiv. & \cmark & & 2 & Contr.\ MLP & Adversarial and covariance debiasing \\
    FairGT~\cite{DBLP:conf/ijcai/LuoHY0024} & IJCAI'24 & Class. & \xmark & & \cmark & & 2 & Hop-token attn & $S$-aware hop aggregation, eigenvector encodings \\
    FairGP~\cite{DBLP:conf/aaai/LuoH0X0025} & AAAI'25 & Class. & \xmark & & & \cmark & -- & Part.\ attn & No attention across partitions \\
    \midrule
    \textbf{FairPGT (ours)} & -- & Clust. & \cmark & \cmark & & \cmark & $\ge2$ & Part.\ attn & Routing tokens, nCGAR attention self-distillation \\
    \bottomrule
  \end{tabular}}
\end{table*}
 
\subsection{Deep graph clustering and graph transformers}
Classical community detection optimizes structural criteria such as modularity~\citep{newman2006modularity,fortunato2016community}. Building on this, deep graph clustering models combine representation learning with clustering, through graph autoencoders~\citep{DBLP:conf/cikm/WangPLZJ17,DBLP:conf/www/Bo0SZL020}, differentiable modularity~\citep{DBLP:journals/jmlr/TsitsulinPPM23}, differentiable pooling~\citep{DBLP:conf/nips/DiffPool18,DBLP:conf/icml/BianchiGA20}, contrastive objectives~\citep{DBLP:conf/aaai/YangLZWTZLFZ23,DBLP:journals/tnn/LiuYZLWLTL24}, or cluster dilation and shrinkage on large graphs~\citep{DBLP:conf/icml/dinknet23}. These methods, however, do not consider protected attributes, and their encoders are GNNs or MLPs. Graph transformers, in turn, replace fixed neighborhood aggregation by attention over sets of nodes, using structural or positional encodings to inject graph information~\citep{kreuzer2021rethinking,DBLP:conf/nips/RampasekGDLWB22,DBLP:conf/icml/ChenOB22}, and tokenized designs scale them to large graphs~\citep{DBLP:journals/tbd/ChenLGLH25}. For attributed graph clustering in particular, the work of \citep{DBLP:journals/eswa/ZhangZSRJ25} fuses a graph-transformer autoencoder with a feature autoencoder whose embeddings are adversarially matched to a prior distribution, and train the fused embeddings with a DEC-style self-supervised objective; its adversarial component regularizes the embedding distribution and does not involve protected attributes. In short, these architectures are designed for clustering or prediction quality alone, and how attention carries group information into clusters has not yet been studied.
 
\subsection{Fair clustering}
Early fair clustering enforces demographic balance in each cluster through fairlets~\citep{chierichetti2017fair} or constrained $k$-clustering~\citep{DBLP:conf/nips/BeraCFN19}, while related formulations constrain the group membership of the selected centers instead~\citep{DBLP:conf/icml/HotegniMV23}. Fair spectral clustering carries the same idea to graphs by adding group-proportionality constraints to the spectral relaxation~\citep{kleindessner2019fair,wang2023scalable,DBLP:journals/pr/FengZQPY26}. Beyond groups, individual fairness requires similar individuals to be treated similarly~\citep{DBLP:conf/innovations/DworkHPRZ12,DBLP:conf/nips/BechavodJW20}, and has been formulated for spectral clustering through an individual-level constraint encoded by a representation graph~\citep{DBLP:conf/nips/GuptaD22}. Along the same lines, our previous factorization-based models regularize the cluster memberships for statistical parity~\citep{DBLP:conf/bigdataconf/GhodsiSQ0N25} and for contrastive individual fairness~\citep{DBLP:conf/pakdd/GhodsiSN24}. In community detection, finally, fairness has been defined through the modularity of each group~\citep{DBLP:conf/www/GkartziosPT25} and through the balance of group proportions combined with modularity~\citep{mouflon2025}. What all of these methods share is that fairness is imposed on the clustering itself, through constraints, pre- or post-processing, or regularization of the assignments, while the representation is either the input data or a spectral or factorization embedding of it. Consequently, none of them learns an unbiased message-passing or attention encoder, and bias introduced by such an encoder lies outside their scope.
 
\subsection{Fairness in deep graph learning}
Fair graph learning has mainly targeted node classification. Among the first approaches, FairGNN debiases representations adversarially~\citep{DBLP:conf/wsdm/DaiW21}, NIFTY uses counterfactual augmentation~\citep{DBLP:conf/uai/AgarwalLZ21}, and EDITS edits the graph and node features before training~\citep{DBLP:conf/www/DongLJL22}. Analyses of message passing have since shown that aggregation over neighborhoods dominated by one protected group can amplify group information in node representations~\citep{DBLP:journals/corr/fmp,DBLP:conf/log/LinKCT23}. Accordingly, several methods intervene inside the model rather than on its outputs: FairGAT designs fairness-aware attention for graph attention networks~\citep{DBLP:journals/tkdd/KoseS24}, FMP adds a demographic-parity term to an optimization view of message passing~\citep{DBLP:journals/corr/fmp}, FS-GNN jointly sparsifies edges and weights under a fairness penalty~\citep{DBLP:journals/ijon/ZhaoHLYPFP25}, and FairNCI fuses estimated counterfactual-neighbor representations into each message-passing layer~\citep{DBLP:journals/pr/ChenWCLLZ26}. Fairness has also been treated jointly with privacy, by combining ranking-based individual fairness with adversarial protection against protected-attribute inference~\citep{DBLP:journals/nn/WangLGBCCL26}. More recently, two graph transformers have addressed fairness, again for node classification: FairGT builds fairness-aware structural and multi-hop feature encodings~\citep{DBLP:conf/ijcai/LuoHY0024}, and FairGP partitions the graph and removes attention between partitions~\citep{DBLP:conf/aaai/LuoH0X0025}. Notably, all of these methods are trained for node classification, and some, such as FS-GNN and FairNCI, even use class labels in the fairness mechanism itself.
 
Deep fair clustering, in contrast, debiases representations or assignments without labels. In the vision domain, DFC filters protected attributes out of learned representations~\citep{DBLP:conf/cvpr/LiZL20}, and for multi-view data, AFMVC removes them from fused representations with a gradient-reversal discriminator while aligning view-specific soft assignments to a consensus $k$-means target~\citep{DBLP:journals/pr/JiangZHLHC26}. On graphs, FairGC combines cross-view contrastive learning with adversarial and covariance-based debiasing for group and individual fairness~\citep{DBLP:conf/aaai/ZhangWRWYMMLJ26}, and modularity-fair deep community detection adds group-modularity terms to the DMoN objective~\citep{DBLP:conf/icdm/GkartziosPT25}. Yet none of their encoders uses attention. Taken together, attention-level fairness has so far been studied only in supervised graph attention networks and graph transformers, whereas fair clustering corrects the representations and assignments of attention-free encoders; the two lines have not met.

\paragraph{\textbf{Positioning our work}} FairPGT brings fairness into the attention of a graph transformer for unsupervised clustering, a combination that, to our knowledge, no previous method addresses (\autoref{tab:comparison}). At its core is a new measure of attention bias, the normalized cross-group attention ratio (nCGAR). Unlike a raw count of attention mass, nCGAR compares cross-group and same-group attention against what the group composition of each partition alone would produce. It therefore separates a genuine preference for one's own group from the mere fact that a larger group receives more attention, and it reads directly: $1$ means group-neutral attention, values below $1$ mean same-group preference, and values above $1$ mean over-correction toward the other group. Because it needs no labels, nCGAR serves both as a training signal, through attention self-distillation, and as an audit of any trained encoder. As in FairGP, attention is computed within METIS partitions; unlike FairGP, however, information still flows across partitions through routing tokens instead of being cut. Above all, FairPGT is the first method to treat fairness at both levels at once: nCGAR self-distillation keeps the encoder group-neutral, while a mutual-information term makes the clusters independent of the protected attribute, and a graph-aware self-training target keeps them aligned with the community structure. As our experiments show, the two fairness levels are complementary, since each alone leaves the other unsolved. Throughout, the protected attribute enters only the training objective and is never an input to the transformer.

\section{FairPGT: Partitioned Graph Transformers for Dual-Level Fair Clustering} %
\label{sec:method}

\textbf{Problem formulation.}
Let $\mathcal{G} = (\mathcal{V}, \mathcal{E})$ be an undirected attributed graph with $n = |\mathcal{V}|$ nodes, adjacency matrix $A \in \{0,1\}^{n \times n}$, node degrees $d_i = \sum_j A_{ij}$, and normalized Laplacian $L_{\mathrm{sym}} = I - \bar A$, where $\bar A_{ij} = A_{ij}/\sqrt{d_i d_j}$. Each node $i$ carries a feature vector $x_i \in \mathbb{R}^d$, collected as $X \in \mathbb{R}^{n \times d}$, and a protected attribute $s_i \in \{1,\dots,G\}$, collected as $S \in \{1,\dots,G\}^n$. Given $(A, X, S)$, our goal is to learn a model $f_\theta$ that maps $(A,X)$ to soft cluster assignments $C = f_\theta(A,X) \in \mathbb{R}^{n \times K}$, with $C_{ik} \geq 0$ and $\sum_k C_{ik} = 1$, where $K$ is the number of clusters, such that the clusters are both structurally coherent and demographically fair. Beyond fair clusters, we also ask that the encoder's attention be balanced across groups, a requirement we make precise through the attention of the model in \autoref{sec:encoder}. We work with soft assignments because they allow clustering and fairness objectives to be optimized jointly by gradient descent, which has proven effective in deep clustering~\citep{xie2016dec,DBLP:journals/jmlr/TsitsulinPPM23}; the final hard labels used for evaluation are then $\hat y_i=\arg\max_k C_{ik}$. Importantly, the protected attributes are not part of the feature matrix $X$: they are provided separately as $S$ and enter only the training objective, through the attention self-distillation term and the group-fairness term (\autoref{sec:encoder} and \autoref{sec:objective}). The trained model therefore computes cluster assignments from $(A,X)$ alone.

\begin{figure*}[t]
    \centering
    \includegraphics[width=1\linewidth]{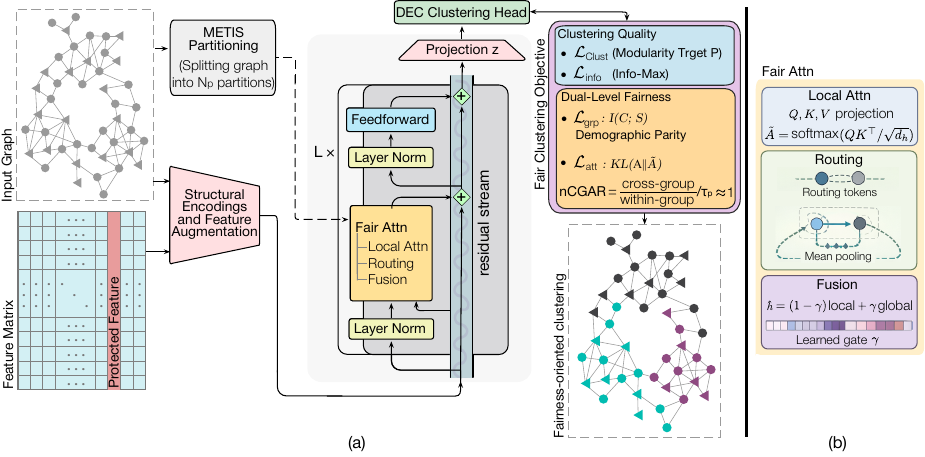}
    \caption{\textbf{FairPGT Pipeline:}
    (a)~Structural encodings augment the node features, partitioned transformer layers produce embeddings, and a DEC-style head outputs soft assignments $C$. The objective combines two quality terms, the graph-aware target $P$ and the information term, with two fairness terms: $\mathcal L_{\mathrm{grp}}$ on the assignments and $\mathcal L_{\mathrm{att}}$, built from the nCGAR target on the local attention. The protected attribute (red column) is not an input to the encoder; it enters only the fairness terms during training. 
    (b)~Fair attention module: local attention within METIS partitions, routing tokens that carry context across partitions, and a learned scalar gate that fuses the two.}
    \label{fig:pipeline}
    \vspace{-4mm}
\end{figure*}

\textbf{Framework overview.}
\autoref{fig:pipeline} illustrates the FairPGT pipeline, with algorithmic details in \autoref{alg:fairpgt_train} and~\autoref{alg:fairpgt_layer}. The framework first augments the node features with structural encodings, then processes them through partitioned transformer layers whose routing tokens carry context across partitions (\autoref{sec:encoder}), and finally produces soft cluster assignments (\autoref{sec:objective}). These assignments are trained toward a graph-aware self-training target, together with an information term that keeps them confident and non-degenerate. On top of this, fairness enters at two levels. At the encoder level, an nCGAR target (\autoref{def:cgar}) rebalances the attention within each partition, and the encoder learns to reproduce it through self-distillation; at the assignment level, a mutual-information term promotes statistical parity across clusters. Notably, both fairness terms belong to the training objective, so the protected attribute is never an input of the model, and inference uses $(A,X)$ alone.

\subsection{FairPGT Encoder}
\label{sec:encoder}
The encoder maps the augmented input to node representations through $L$ transformer layers. Each layer combines three components: (i)~partition-local attention, which keeps the cost of attention manageable; (ii)~routing tokens, which carry context across partitions; and (iii)~a learned gate that fuses local and global information. In addition, during training, the local attention is guided toward group-neutrality by a target built from our normalized cross-group attention ratio (nCGAR). Throughout, $\mathrm{MHA}(\cdot)$ denotes standard multi-head attention, $\mathrm{FFN}(\cdot)$ a position-wise feed-forward network, and $\mathrm{LN}(\cdot)$ layer normalization. We now describe each component in turn.

\textbf{Structural encodings.}
Since attention by itself is blind to graph structure, we first augment the node features with lightweight descriptors derived from the graph. Let $(\lambda_m, u_m)$ denote the eigenpairs of the Laplacian $L_{\mathrm{sym}}$. We compute the normalized degree $\tilde{d}_i = d_i / \max_j d_j$ and the first $b$ nontrivial Laplacian eigenvectors $U = [u_1,\dots,u_b] \in \mathbb{R}^{n \times b}$. The initial token matrix
\begin{equation}
H^{(0)} = [\,X \;\|\; U \;\|\; \tilde{d}\,]
\in \mathbb{R}^{n \times d_0}, \quad d_0 = d + b + 1,
\label{eq:init-tokens}
\end{equation}
thus combines node attributes with spectral position information and degree-based centrality. Before training, each column of $H^{(0)}$ is standardized to zero mean and unit variance, so that attributes, spectral encodings, and degrees enter on a comparable scale. Since eigenvectors are defined only up to sign, we fix the sign of each so that its largest-magnitude entry is positive, and we let the learned input projection absorb the remaining ambiguity for repeated eigenvalues.

\textbf{Partitioned attention with routing.}
Full attention over all nodes costs $\mathcal{O}(n^2)$ per layer, which quickly becomes prohibitive on large graphs. To scale beyond it, we partition the graph into $N_p$ disjoint blocks $\mathcal P = \{\mathcal V_1,\dots,\mathcal V_{N_p}\}$ using METIS~\cite{DBLP:journals/siamsc/KarypisK98}, a widely used multilevel partitioning algorithm that scales to large graphs and keeps densely connected nodes together. We set $N_p=\max(8,\lceil n/2500\rceil)$, so that partitions hold at most about 2{,}500 nodes; the resulting complexity reduction is analyzed in \autoref{sec:complexity}, and \autoref{sec:cost} studies other values of $N_p$. Within each partition $\mathcal V_p$ and for every layer $\ell$, local multi-head attention is then computed from the projected representations $Q_{p}, K_{p}, V_{p} = \mathrm{LN}(H_{p}^{(\ell-1)})W_{\{Q,K,V\}}^{(\ell)}$ as
\begin{equation}
\tilde{\mathcal{A}}_{p}^{(\ell)} = \mathrm{softmax}\left( Q_{p}^{(\ell)}{K_{p}^{(\ell)}}^\top / \sqrt{d_h} \right), \qquad p \in \{1,\dots,N_p\}.
\label{eq:local-attn}
\end{equation}

Attention is computed per head; we omit the head index for readability, and all attention-level quantities below are likewise computed per head. Moreover, the diagonal of $Q_pK_p^\top$ is masked before the softmax, so that a node attends only to the other nodes of its partition, while its own state enters through the residual connection in Eq.~\eqref{eq:gated-sum}.

Partitioning alone, however, limits the receptive field to within-partition nodes and severs the long-range communication that motivates graph transformers in the first place. To restore global context, we introduce one \emph{routing token} per partition, following the inducing-point intuition of Set Transformer~\cite{DBLP:conf/icml/LeeLKKCT19}. Specifically, at layer $\ell$ each partition is summarized by the mean of its normalized states, $r_{p}^{(\ell)} = |\mathcal{V}_{p}|^{-1}\sum_{i\in\mathcal{V}_{p}}\mathrm{LN}(h_i^{(\ell-1)})$. The resulting set of tokens $R^{(\ell)}=[r_1^{(\ell)};\dots;r_{N_p}^{(\ell)}]$ exchanges information through global self-attention with a residual connection, $\hat{R}^{(\ell)}=R^{(\ell)}+\mathrm{MHA}\big(\mathrm{LN}(R^{(\ell)}),$ $\mathrm{LN}(R^{(\ell)}),\mathrm{LN}(R^{(\ell)})\big)$, and each node then absorbs global context through cross-attention to the tokens, $g_i^{(\ell)}=\mathrm{MHA}\big(\mathrm{LN}(h_i^{(\ell-1)}),\hat{R}^{(\ell)}, \hat{R}^{(\ell)}\big)$. Since there are only $N_p$ tokens, this global exchange adds little cost. Finally, local and global outputs are combined through a learned gate, followed by a feed-forward block, each with a residual connection:
\begin{equation}
\hat h_i^{(\ell)} = h_i^{(\ell-1)} + (1{-}\gamma^{(\ell)})\,\big[\tilde{\mathcal{A}}_{p}^{(\ell)}V_{p}^{(\ell)}W_O^{(\ell)}\big]_i
  + \gamma^{(\ell)}\,g_i^{(\ell)},
\qquad
h_i^{(\ell)} = \hat h_i^{(\ell)} + \mathrm{FFN}\big(\mathrm{LN}(\hat h_i^{(\ell)})\big),
\label{eq:gated-sum}
\end{equation}
where $i \in \mathcal V_p$, $W_O^{(\ell)}$ is the output projection, and $\gamma^{(\ell)}=\sigma(\bar{\gamma}^{(\ell)})\in(0,1)$ is a learned scalar that controls the local--global trade-off, so that each layer decides for itself how much global context it needs. \autoref{alg:fairpgt_layer} summarizes one encoder layer. Importantly, the layer does not take the protected attributes as input; it only returns the partition attention matrices, from which the CGAR targets are built during training (\autoref{alg:fairpgt_train}).

\begin{algorithm}[t]
\caption{FairPGT: training and inference. Steps marked $\blacktriangleright$ are the only ones that use the protected attribute $S$; inference uses $(A,X)$ alone.}
\label{alg:fairpgt_train}
\begin{algorithmic}[1]
\Require Graph $A$, features $X$, protected attribute $S$ (training only), number of clusters $K$; weights $\lambda_{\mathrm{grp}},\lambda_{\mathrm{att}},\lambda_{\mathrm{info}}$; CGAR step $\alpha$ and tolerance $\epsilon$; target interval $T_P$; epochs $T$
\Ensure Trained model $f_\theta$; soft assignments $C$ and cluster labels $\hat y$
\Statex \textbf{Preprocessing (once)}
\State Build the input tokens $H^{(0)}$ from features, Laplacian eigenvectors, and degrees, standardized per column \Comment{Eq.~\eqref{eq:init-tokens}}
\State Partition the graph with METIS into $N_p=\max(8,\lceil n/2500\rceil)$ blocks $\mathcal P$
\State $\blacktriangleright$ Record the group composition $\tau_p$ of every partition with at least two groups \Comment{Eq.~\eqref{eq:cgar}}
\State Initialize $\theta$, and place the centers $\{\mu_k\}$ by $k$-means on the initial embeddings
\Statex \textbf{Training}
\For{$t=1,\dots,T$}
  \State \textit{Encode:} $H^{(L)}$ and attention $\{\tilde{\mathcal A}_p^{(\ell)}\}$ from $(H^{(0)},\mathcal P)$ \Comment{\autoref{alg:fairpgt_layer}}
  \State \textit{Assign:} project $z_i$ and compute the soft assignments $C$ \Comment{Eq.~\eqref{eq:dec}}
  \If{$(t-1)\bmod T_P=0$}
    \State \textit{Refresh:} recompute the graph-aware target $P$ and hold it fixed \Comment{Eqs.~\eqref{eq:mod-gain}--\eqref{eq:unified-target}}
  \EndIf
  \State $\blacktriangleright$ \textit{Rebalance:} build the CGAR target $\mathcal A_p^{(\ell)}$ for every layer, head, and mixed partition, held fixed \Comment{Eq.~\eqref{eq:cgar_correction}}
  \State $\blacktriangleright$ \textit{Score:} $\mathcal L\gets\mathcal L_{\mathrm{clust}}+\lambda_{\mathrm{grp}}\mathcal L_{\mathrm{grp}}+\lambda_{\mathrm{att}}\mathcal L_{\mathrm{att}}+\lambda_{\mathrm{info}}\mathcal L_{\mathrm{info}}$ \Comment{Eq.~\eqref{eq:total-loss}}
  \State \textit{Update:} one Adam step on the encoder, the gates, $W_z$, and the centers
\EndFor
\Statex \textbf{Inference}
\State Encode and assign as in training: $C\gets f_\theta(A,X)$
\State $\hat y_i\gets\arg\max_k C_{ik}$ for every node $i$
\State \Return $f_\theta$, $C$, $\hat y$
\end{algorithmic}
\end{algorithm}

\textbf{nCGAR-guided fair attention.}
While partitioning limits how far bias can propagate, attention within a partition can still concentrate on same-group nodes. This matters because, when a node's representation is aggregated mostly from nodes of its own protected group, group membership becomes easier to recover from the representation, and clusters formed on such representations tend to follow group boundaries. The same effect has been shown for message passing on graphs with protected-attribute homophily~\citep{DBLP:journals/corr/fmp,DBLP:conf/log/LinKCT23}, and fairness-aware graph transformers have likewise shown that attention can carry protected information~\citep{DBLP:conf/ijcai/LuoHY0024,DBLP:conf/aaai/LuoH0X0025}. To measure this effect, we introduce the \emph{Cross-Group Attention Ratio} (CGAR), which compares the attention mass on cross-group and within-group node pairs inside a partition, and its normalized form, nCGAR, which relates this ratio to what the group composition of the partition alone would produce. With this measure in hand, FairPGT addresses group dependence directly in the encoder, complementing the assignment-level fairness term of \autoref{sec:objective}.

\begin{definition}[Cross-Group Attention Ratio]
\label{def:cgar}
For partition $\mathcal{V}_p$ at layer $\ell$ with attention matrix $\tilde{\mathcal{A}}_p^{(\ell)}$, the CGAR and its normalized form nCGAR are
\begin{equation}
\mathrm{CGAR}_p^{(\ell)} =
\frac{\sum_{i,j\in\mathcal V_p:\, s_i \neq s_j} \tilde{\mathcal A}_{ij}^{(\ell)}}
     {\sum_{i\neq j\in\mathcal V_p:\, s_i = s_j} \tilde{\mathcal A}_{ij}^{(\ell)}},
\qquad
\mathrm{nCGAR}_p^{(\ell)} = \frac{\mathrm{CGAR}_p^{(\ell)}}{\tau_p},
\qquad
\tau_p = \frac{M_p^{\times}}{M_p^{=}} = \frac{n_p^2-\sum_g n_{pg}^2}{\sum_g n_{pg}(n_{pg}-1)},
\label{eq:cgar}
\end{equation}
where $n_{pg}$ is the number of nodes of group $g$ in $\mathcal V_p$, $n_p=\sum_g n_{pg}$, and $M_p^{\times}$ and $M_p^{=}$ are the numbers of ordered cross-group and within-group pairs of distinct nodes in $\mathcal V_p$. CGAR is defined for partitions that contain at least two groups and at least one within-group pair.
\end{definition}

To see why the normalization is needed, consider group-agnostic attention, for example uniform attention within the partition. Each cross-group pair then receives on average as much attention as each within-group pair, so $\mathrm{nCGAR}_p=1$, whereas the raw ratio equals $\tau_p$, which is close to $G-1$ for $G$ equally sized groups and depends on the composition of the partition otherwise. Accordingly, $\mathrm{nCGAR}_p<1$ indicates \emph{group-homophilous} attention, concentrated on same-group pairs beyond what the composition explains, and $\mathrm{nCGAR}_p>1$ indicates \emph{group-heterophilous} attention, an over-correction toward the other group. Because it is normalized, nCGAR can be compared across partitions, layers, heads, and datasets, and because it needs no labels, it serves both as a training signal and as an audit of a trained encoder (\autoref{sec:sbm}). Naturally, nCGAR is defined only for partitions that contain at least two groups.
Moreover, nCGAR pools the attention of all nodes in the partition: it measures whether cross-group and within-group pairs receive balanced attention overall. A value of 1 can therefore also arise when one group attends across groups more than the composition explains and another group less, a case we examine in \autoref{sec:sbm}. Throughout, we call attention with nCGAR $=1$
group-neutral in this aggregate sense.

We turn nCGAR into a training signal through a simple correction. Whenever $\mathrm{nCGAR}_p^{(\ell)} < 1-\epsilon$, we multiplicatively boost the cross-group attention entries; conversely, whenever $\mathrm{nCGAR}_p^{(\ell)} > 1+\epsilon$, we boost the within-group entries. The result serves as a target for the encoder's attention:
\begin{equation}
\mathcal{A}_{ij}^{(\ell)} \propto
\begin{cases}
\tilde{\mathcal{A}}_{ij}^{(\ell)}(1+\alpha),
  & \text{if }
    \begin{cases} s_i \neq s_j, \text{nCGAR}_p < 1-\epsilon \\
                  s_i = s_j, \text{nCGAR}_p > 1+\epsilon
    \end{cases} \\
\tilde{\mathcal{A}}_{ij}^{(\ell)},
  & \text{otherwise},
\end{cases}
\label{eq:cgar_correction}
\end{equation}
after which row renormalization restores stochasticity. As with the DEC target $P$ in \autoref{sec:objective}, the target $\mathcal A$ is recomputed from the current attention at every training step and treated as a constant, i.e., no gradient flows through it. The encoder's attention is then aligned with this target through
\begin{equation}
\mathcal{L}_{\mathrm{att}} = \frac{1}{|\mathcal I|}\sum\nolimits_{(\ell,p,i)\in\mathcal I}
\mathrm{KL}\big(\mathcal A_{p,i}^{(\ell)} \,\big\|\, \tilde{\mathcal A}_{p,i}^{(\ell)}\big),
\label{eq:att-align}
\end{equation}
where $\mathcal A_{p,i}^{(\ell)}$ is row $i$ of the attention matrix of partition $\mathcal V_p$ at layer $\ell$, and $\mathcal I$ collects all rows over layers, partitions with at least two groups, and heads. In the spirit of self-distillation~\citep{DBLP:conf/iccv/KimJYH21}, the encoder thus learns to produce group-neutral attention by itself, by distilling a rebalanced version of its own attention. The tolerance $\epsilon$ leaves small deviations from the composition baseline unpenalized. On the theoretical side, we prove in \autoref{thm:cgar} that the target always moves nCGAR toward the tolerance band $[1{-}\epsilon, 1{+}\epsilon]$, and in \autoref{prop:align} that minimizing $\mathcal{L}_{\mathrm{att}}$ moves the encoder's own attention in the same direction, with $\mathcal{L}_{\mathrm{att}}=0$ once every partition lies inside the band. Finally, since the corrected attention appears only in the loss, the forward pass in Eq.~\eqref{eq:gated-sum} uses $\tilde{\mathcal A}$ during both training and inference, and the protected attributes are never required at inference.

\subsection{Fairness-aware clustering}
\label{sec:objective}
\textbf{Soft assignments.} On top of the encoder, a clustering head turns node representations into soft cluster assignments. Let $H^{(L)}$ denote the final encoder output and $z_i = \mathrm{LN}(W_z h_i^{(L)}) \in \mathbb{R}^{d_z}$ its low-dimensional projection. We adopt the DEC clustering head~\cite{xie2016dec}, which has become a standard component of deep graph clustering thanks to its differentiability and its compatibility with self-training objectives~\citep{DBLP:conf/www/Bo0SZL020,DBLP:journals/jmlr/TsitsulinPPM23}. Given learnable centers $\{\mu_k\}_{k=1}^K$, initialized by $k$-means on the initial embeddings, the row-stochastic soft assignment matrix $C \in \mathbb{R}^{n \times K}$ is computed with a Student's $t$ kernel as
\begin{equation}
C_{ik} = \delta_{ik}^{-1} / \textstyle\sum_{k'} \delta_{ik'}^{-1}, \qquad \delta_{ik} = 1 + \|z_i - \mu_k\|_2^2.
\label{eq:dec}
\end{equation}

\textbf{Graph-aware target.} Standard DEC sharpens the assignments using only $C$ and thus ignores the graph. To bring the community structure into the target, we first score each node--cluster pair by its modularity gain,
\begin{equation}
M_{ik} = \sum\nolimits_{j \in \mathcal N(i)} C_{jk}
-\frac{d_i}{2|\mathcal E|}\sum\nolimits_{j=1}^n d_j C_{jk}, \qquad
\mathcal N(i)=\{j:A_{ij}=1\},
\label{eq:mod-gain}
\end{equation}
where $|\mathcal E|$ is the number of edges and $\mathcal N(i)$ the neighborhood of node $i$. The first term measures how strongly the neighbors of $i$ belong to cluster $k$, and the second subtracts what a degree-preserving random graph would produce, so that $M_{ik} > 0$ means that the neighbors of node $i$ favor cluster $k$ beyond random connectivity. To keep these scores on a bounded scale, we center them and apply a parameter-free nonlinearity,
\begin{equation}
\tilde{M}_{ik} = \tanh(M_{ik} - \bar{M}_i), \qquad \bar{M}_i = \tfrac{1}{K} \sum\nolimits_{r=1}^K M_{ir}.
\label{eq:stable-score}
\end{equation}
Under exact row-stochasticity of $C$ the gains already satisfy $\sum_k M_{ik}=0$, so the centering merely guards against finite-precision drift during training. The graph-aware target then combines DEC confidence with this structural guidance,
\begin{equation}
P_{ik} =
\frac{(C_{ik}^2/f_k)\exp(\tilde M_{ik})}
     {\sum_{k'}(C_{ik'}^2/f_{k'})\exp(\tilde M_{ik'})},
\qquad
f_k = \sum\nolimits_i C_{ik},
\label{eq:unified-target}
\end{equation}
and the clustering loss is $\mathcal{L}_{\mathrm{clust}}=\mathrm{KL}(P\|C)$. Following DEC, $P$ is held fixed during back-propagation and recomputed every $T_P$ training steps. Intuitively, when the graph structure is informative, $\tilde M_{ik}$ sharpens cluster preferences along local modularity gains; when it is weak or homogeneous, the target reduces to standard DEC. Moreover, because $\tilde M_{ik}$ is bounded, the structural guidance can tip close calls but cannot overturn confident DEC preferences: \autoref{thm:target} shows that $P$ always stays within a factor $e^{2}$ of the DEC target.

\textbf{Group fairness.} Next, we turn to the fairness of the assignments. Let $p^*(g)=|\{i:s_i=g\}|/n$ be the global proportion of group $g$, and let $\hat p_k(g)=\sum_{i:s_i=g} C_{ik}/\sum_i C_{ik}$ be its soft proportion in cluster $k$. We penalize deviations from statistical parity by the mutual information between cluster and group under the soft assignments,
\begin{equation}
\mathcal{L}_{\mathrm{grp}} = I(C;S) = \sum\nolimits_{k=1}^K \bar c_k\,\mathrm{KL}(\hat p_k \,\|\, p^*),
\qquad \bar c_k=\frac{1}{n}\sum\nolimits_{i=1}^n C_{ik},
\label{eq:grp}
\end{equation}
which is zero exactly when every cluster has the global group proportions. Weighting each cluster by its expected size $\bar c_k$ keeps small clusters from dominating the penalty. A small constant is added to all proportions for numerical stability.

\textbf{Informative assignments.} Statistical parity alone, however, admits degenerate solutions. For example, if all nodes share the same assignment vector, every cluster inherits exactly the global group proportions, so $\mathcal{L}_{\mathrm{grp}}=0$, although the clustering carries no structure at all. To exclude such solutions, we additionally require the assignments to be informative about the nodes. Inspired by the information-maximization view of discriminative clustering~\citep{DBLP:conf/nips/GomesKP10}, we minimize the negative mutual information between a node and its cluster under the soft assignments,
\begin{equation}
\mathcal{L}_{\mathrm{info}} = \frac{1}{n}\sum\nolimits_{i=1}^n H(C_i) - H(\bar c),
\label{eq:info}
\end{equation}
where $H(p)=-\sum_{k=1}^{K}p_k\log p_k$ is the Shannon entropy (natural logarithm) of a distribution $p$ over the $K$ clusters, $C_i$ is the assignment vector of node $i$, and $\bar c=(\bar c_1,\dots,\bar c_K)$ collects the expected cluster proportions of Eq.~\eqref{eq:grp}. The two terms pull in complementary directions: the first is small when each node is assigned confidently to one cluster, and the second, which enters with a negative sign, is large when the clusters have similar sizes. Since the entropy is concave, $\mathcal{L}_{\mathrm{info}}$ always lies between $-\log K$ and $0$. It reaches its minimum $-\log K$ only when every node is assigned with certainty and all clusters have equal size, and its maximum $0$ exactly when all nodes share the same assignment vector, that is, precisely for the degenerate solutions that satisfy $\mathcal{L}_{\mathrm{grp}}$ trivially. In short, $\mathcal{L}_{\mathrm{clust}}$ and $\mathcal{L}_{\mathrm{info}}$ together serve clustering quality: the former aligns the clusters with the community structure, and the latter keeps the assignments confident and the clusters non-degenerate. Neither involves the protected attribute; fairness is the role of $\mathcal{L}_{\mathrm{grp}}$, the mutual information between clusters and groups, and of $\mathcal{L}_{\mathrm{att}}$.

\textbf{Final objective.} Putting the pieces together, FairPGT minimizes
\begin{equation}
\mathcal{L}_{\text{FairPGT}} = \mathcal{L}_{\mathrm{clust}}
+ \lambda_{\mathrm{grp}}\,\mathcal{L}_{\mathrm{grp}}
+ \lambda_{\mathrm{att}}\,\mathcal{L}_{\mathrm{att}}
+ \lambda_{\mathrm{info}}\,\mathcal{L}_{\mathrm{info}},
\label{eq:total-loss}
\end{equation}
in which two terms serve clustering quality and two serve fairness: $\mathcal{L}_{\mathrm{clust}}$ aligns the clusters with the community structure and $\mathcal{L}_{\mathrm{info}}$ keeps them informative, while $\mathcal{L}_{\mathrm{grp}}$ makes the clusters independent of the protected attribute and $\mathcal{L}_{\mathrm{att}}$ keeps the encoder's attention balanced across groups. Accordingly, $\lambda_{\mathrm{grp}}$ and $\lambda_{\mathrm{att}}$ control the fairness--utility trade-off at the assignment and attention levels, and $\lambda_{\mathrm{info}}$ sets the strength of the information term; in addition, $\alpha$ and $\epsilon$ set the step size and tolerance of the CGAR target, and the gates $\gamma^{(\ell)}$ are learned end-to-end.

\textbf{Training and inference.} \autoref{alg:fairpgt_train} brings all components together. Structural encodings, partitions, and group compositions are computed once, before training. At each training step, the encoder computes its attention and the soft assignments from $(A,X)$ alone; the CGAR target $\mathcal A$ and the graph-aware target $P$ are then built from these outputs and held fixed, and all parameters are updated on Eq.~\eqref{eq:total-loss}. Notably, the protected attributes enter at only two places, the CGAR target and $\mathcal L_{\mathrm{grp}}$, both of which belong to the training objective. At inference, the trained model therefore computes $C$ from $(A,X)$ alone. One encoder layer is detailed in \autoref{alg:fairpgt_layer}.

\begin{algorithm}[t]
\caption{FairPGT encoder layer $\ell$. The layer never receives the protected attribute; it returns its attention matrices so that the CGAR targets can be built during training.}
\label{alg:fairpgt_layer}
\begin{algorithmic}[1]
\Require States $H^{(\ell-1)}\in\mathbb R^{n\times D}$, partitions $\mathcal P=\{\mathcal V_1,\dots,\mathcal V_{N_p}\}$
\Ensure States $H^{(\ell)}$ and partition attention matrices $\{\tilde{\mathcal A}_p^{(\ell)}\}$
\State Normalize the states: $\bar H\gets\mathrm{LN}(H^{(\ell-1)})$ \Comment{shared by both branches}
\Statex \textbf{Local branch: attention within partitions}
\For{each partition $\mathcal V_p\in\mathcal P$ (in parallel)}
  \State Compute the self-masked attention $\tilde{\mathcal A}_p^{(\ell)}$ from $\bar H_p$ \Comment{Eq.~\eqref{eq:local-attn}, per head}
  \State Aggregate the values: $O_p\gets\tilde{\mathcal A}_p^{(\ell)}V_pW_O^{(\ell)}$
\EndFor
\Statex \textbf{Global branch: routing tokens}
\State Summarize each partition by a token $r_p$, the mean of its normalized states
\State Let the tokens exchange information by self-attention, giving $\hat R$
\State Broadcast: every node attends to the tokens, giving its global context $g_i$
\Statex \textbf{Fusion}
\State Mix local and global outputs with the learned gate $\gamma^{(\ell)}$, then apply the feed-forward block \Comment{Eq.~\eqref{eq:gated-sum}}
\State \Return $H^{(\ell)}$, $\{\tilde{\mathcal A}_p^{(\ell)}\}$
\end{algorithmic}
\end{algorithm}

\subsection{Theoretical Guarantees}
\label{sec:theory}
We now show that the two targets that drive training behave as intended: the CGAR target moves attention toward neutrality without overshooting, its loss pulls the encoder's own attention in the same direction, and the graph-aware target refines the DEC target without overriding it. All proofs are given in \autoref{sec:proofs}.

\begin{theorem}[CGAR target]
\label{thm:cgar}
Consider one head and one partition $\mathcal V_p$ with at least two groups, attention $\tilde{\mathcal A}$, and target $\mathcal A$ from Eq.~\eqref{eq:cgar_correction}. Let $q_i=\sum_{j:s_j\neq s_i}\tilde{\mathcal A}_{ij}$ and $q_i'=\sum_{j:s_j\neq s_i}\mathcal A_{ij}$ be the cross-group mass of row $i$ before and after the correction.\par
\textup{(i)}~If $\mathrm{nCGAR}(\tilde{\mathcal A})<1-\epsilon$, then $q_i'=(1+\alpha)q_i/(1+\alpha q_i)\ge q_i$ for every row, with strict inequality whenever $q_i<1$, and $\mathrm{nCGAR}(\tilde{\mathcal A})<\mathrm{nCGAR}(\mathcal A)\le(1+\alpha)\,\mathrm{nCGAR}(\tilde{\mathcal A})$.\par
\textup{(ii)}~If $\mathrm{nCGAR}(\tilde{\mathcal A})>1+\epsilon$, then $q_i'=q_i/(1+\alpha(1-q_i))\le q_i$ for every row, with strict inequality whenever $q_i<1$, and $\mathrm{nCGAR}(\tilde{\mathcal A})/(1+\alpha)\le\mathrm{nCGAR}(\mathcal A)<\mathrm{nCGAR}(\tilde{\mathcal A})$.\par
\textup{(iii)}~If $\alpha\le 2\epsilon/(1-\epsilon)$, the target does not overshoot: whenever the correction is active, $\mathrm{nCGAR}(\mathcal A)$ is strictly closer to the band $[1-\epsilon,1+\epsilon]$ than $\mathrm{nCGAR}(\tilde{\mathcal A})$.
\end{theorem}
Parts (i) and (ii) show that one correction step always moves nCGAR in the right direction, by at most a factor $1+\alpha$. Part (iii) is needed because a large $\alpha$ could push nCGAR from just below the band to far above it; the condition holds for any $\alpha\le\epsilon$, including our setting $\alpha=\epsilon=0.05$.

The theorem concerns the target, whereas training changes the encoder. The next result connects the two.

\begin{proposition}[Attention alignment]
\label{prop:align}
\textup{(i)}~$\mathcal{L}_{\mathrm{att}}=0$ if and only if $\mathrm{nCGAR}\in[1-\epsilon,1+\epsilon]$ for every layer, head, and partition with at least two groups.\par
\textup{(ii)}~Let $\tilde{\mathcal A}_i=\mathrm{softmax}(a_i)$ be a row with logits $a_i$, and $\ell_i=\mathrm{KL}(\mathcal A_i\,\|\,\tilde{\mathcal A}_i)$ with the target $\mathcal A_i$ held fixed. Then $\nabla_{a_i}\ell_i=\tilde{\mathcal A}_i-\mathcal A_i$, and if the target boosts cross-group entries, the gradient step $a_i-\eta\nabla_{a_i}\ell_i$ changes $q_i$ at rate
\[
\frac{dq_i}{d\eta}\Big|_{\eta=0}
=\frac{\alpha}{1+\alpha q_i}\Big[(1-q_i)^2\!\!\sum_{j:s_j\neq s_i}\!\tilde{\mathcal A}_{ij}^2+q_i^2\!\!\sum_{j\neq i:\,s_j=s_i}\!\tilde{\mathcal A}_{ij}^2\Big]\ge 0,
\]
with strict inequality whenever $q_i<1$; if the target boosts within-group entries, the rate is the same bracket times $-\alpha/(1+\alpha(1-q_i))\le 0$.
\end{proposition}
In words, part (i) shows that the loss vanishes exactly when the encoder's attention is already within the band, and part (ii) shows that descending $\mathcal{L}_{\mathrm{att}}$ moves the cross-group mass of each row in the same direction as the target. Part (ii) holds for the logits of each row; since all rows share the projections $W_Q$ and $W_K$, a parameter update need not move every row at once. We therefore measure the nCGAR of the trained encoder directly in \autoref{sec:experiments}.

Finally, we turn to the clustering side and characterize how the graph-aware target relates to the standard DEC target.

\begin{theorem}[Graph-aware target]
\label{thm:target}
Let $P^{\mathrm{DEC}}_{ik}=(C_{ik}^2/f_k)/\sum_{k'}(C_{ik'}^2/f_{k'})$ be the standard DEC target. For every node $i$ and clusters $k,k'$, the target $P$ of Eq.~\eqref{eq:unified-target} satisfies:\par
\textup{(i)}~\emph{Bounded modulation:} $e^{-2}P^{\mathrm{DEC}}_{ik}<P_{ik}<e^{2}P^{\mathrm{DEC}}_{ik}$.\par
\textup{(ii)}~\emph{Agreement:} $P_{ik}/P_{ik'}=\big(P^{\mathrm{DEC}}_{ik}/P^{\mathrm{DEC}}_{ik'}\big)\exp\big(\tilde M_{ik}-\tilde M_{ik'}\big)$. Hence, if $M_{ik}>M_{ik'}$ and $P^{\mathrm{DEC}}_{ik}\ge P^{\mathrm{DEC}}_{ik'}$, then $P_{ik}>P_{ik'}$; and $P$ can reverse a DEC preference for $k$ over $k'$ only if $P^{\mathrm{DEC}}_{ik}/P^{\mathrm{DEC}}_{ik'}<e^{2}$ and $M_{ik'}>M_{ik}$.\par
\textup{(iii)}~\emph{Reduction to DEC:} if $M_{ik}$ is the same for all $k$, then $P_i=P^{\mathrm{DEC}}_i$.
\end{theorem}
The graph structure therefore reweights the DEC target but cannot override it: it reinforces the DEC preference when the two agree, it can overturn a weak preference in favor of the cluster preferred by the node's neighbors, and it cannot reverse a confident one, with odds of at least $e^2\approx7.4$. When the modularity gains of a node do not distinguish between clusters, the target is exactly the DEC target.

\subsection{Computational Complexity}
\label{sec:complexity}
Having established what the training targets do, we turn to what the model costs. Throughout, $D=Hd_h$ denotes the width of the encoder with $H$ heads, and $d_0$ the input dimension of Eq.~\eqref{eq:init-tokens} and $d_z$ the dimension of the clustering projection.

\begin{proposition}[Complexity of FairPGT]
\label{thm:complexity}
Let the encoder have $L$ layers of width $D$, and let the $N_p$ partitions be balanced.\par
\textup{(i)}~One encoder layer costs $\mathcal{O}\big(n^2D/N_p+nN_pD+N_p^2D+nD^2\big)$ time.\par
\textup{(ii)}~One full-batch training step costs $\mathcal{O}\big(L(n^2D/N_p+nN_pD+N_p^2D+nD^2)+nd_0D+nDd_z+nKd_z+|\mathcal E|K\big)$ time and $\mathcal{O}\big(LHn^2/N_p+LHnN_p+LnD+|\mathcal E|+nK\big)$ memory for activations.\par
\textup{(iii)}~Inference has the time cost of (ii) without the $|\mathcal E|K$ term and needs $\mathcal{O}(Hn^2/N_p+nD)$ memory, since attention matrices can be released layer by layer.\par
For $N_p=\Theta(\sqrt n)$, one layer costs $\mathcal{O}(n^{3/2}D+nD^2)$ time, compared to $\mathcal{O}(n^2D+nD^2)$ for global attention.
\end{proposition}
In (i), the four terms are local attention within partitions of size $n_p=n/N_p$, the broadcast from the $N_p$ routing tokens to all $n$ nodes, self-attention among the routing tokens, and the linear projections and feed-forward network, which any transformer shares. In (ii), the remaining terms are the input embedding, the clustering projection, the DEC assignments (Eq.~\eqref{eq:dec}), and the graph-aware target (Eq.~\eqref{eq:unified-target}), which needs $\mathcal{O}(|\mathcal E|K)$ for sparse neighbor aggregation and never forms dense $n\times n$ matrices. In contrast, the CGAR target (Eq.~\eqref{eq:cgar_correction}) and $\mathcal{L}_{\mathrm{att}}$ (Eq.~\eqref{eq:att-align}) reuse the partition attention matrices that the forward pass has already computed, while $\mathcal{L}_{\mathrm{grp}}$ and $\mathcal{L}_{\mathrm{info}}$ cost $\mathcal{O}(nK)$; none of them changes the order of the cost, and none is computed at inference. METIS partitioning and the spectral encodings are computed once, before training.

In practice, our default $N_p=\max(8,\lceil n/2500\rceil)$ keeps partitions at about $n_p\le2{,}500$ nodes, so the local-attention term becomes $\mathcal{O}(nn_pD)$, linear in $n$. The broadcast term $nN_pD=n^2D/n_p$ still grows quadratically, but with the small factor $1/n_p$: on Pokec-n ($N_p=27$) it is about 90 times smaller than the local-attention term. \autoref{fig:scaling} confirms that memory follows the local-attention term: peak memory falls close to $1/N_p$, until the per-node terms dominate, from 11.9\,GB to 0.75\,GB on LastFM as $N_p$ grows from 1 to 64, while the time per epoch falls until fixed per-layer costs dominate.

\subsection{Proofs of Theoretical Claims}
\label{sec:proofs}
 
\subsubsection{Proof of \autoref{thm:cgar} (CGAR Target)}
 
\paragraph{Setup}
Fix a layer, a head, and a partition $\mathcal V_p$ with $n_p$ nodes and at least two groups; $s_i$ may take any number of values. Let $\tilde{\mathcal A}\in\mathbb R^{n_p\times n_p}$ be the row-stochastic attention of this partition, whose diagonal is masked, and write $j\in c(i)$ if $s_j\neq s_i$ and $j\in w(i)$ if $s_j=s_i$ and $j\neq i$. The cross-group mass of row $i$ is $q_i=\sum_{j\in c(i)}\tilde{\mathcal A}_{ij}$, and its within-group mass is $1-q_i$. Softmax entries outside the diagonal are positive and $c(i)\neq\emptyset$, so $q_i\in(0,1]$, with $q_i=1$ exactly when node $i$ has no other node of its group in the partition. Such rows are left unchanged by the correction, since all their nonzero entries are scaled by the same factor. Because CGAR is defined only when the partition contains within-group pairs, at least one row has $q_i\in(0,1)$. Summing over rows,
\begin{equation}
\mathrm{CGAR}(\tilde{\mathcal A})=\frac{\sum_i q_i}{\sum_i(1-q_i)},\qquad
\mathrm{nCGAR}(\tilde{\mathcal A})=\frac{\mathrm{CGAR}(\tilde{\mathcal A})}{\tau_p}.
\label{eq:cgar-rows}
\end{equation}
Since $\tau_p>0$ depends only on the group composition of $\mathcal V_p$, it suffices to prove all statements for CGAR.
 
\begin{proof}
\textbf{(i) Correction from below.} The target multiplies the cross-group entries of each row by $1+\alpha$ and renormalizes by $Z_i=(1-q_i)+(1+\alpha)q_i=1+\alpha q_i$. Hence
\[
q_i'=\frac{(1+\alpha)q_i}{1+\alpha q_i},\qquad
q_i'-q_i=\frac{\alpha q_i(1-q_i)}{1+\alpha q_i}\ge 0.
\]
with equality only for rows with $q_i=1$. Since at least one row has $q_i<1$, $\sum_i q_i'>\sum_i q_i$; and because $\sum_i(1-q_i)=n_p-\sum_i q_i$ and $Q\mapsto Q/(n_p-Q)$ is strictly increasing on $[0,n_p)$, this gives $\mathrm{CGAR}(\mathcal A)>\mathrm{CGAR}(\tilde{\mathcal A})$.
 
For the upper bound, let $u_i=1/Z_i$. Then $q_i'=(1+\alpha)q_iu_i$ and $1-q_i'=(1-q_i)u_i$, so
\[
\mathrm{CGAR}(\mathcal A)=(1+\alpha)\,\frac{\sum_i q_iu_i}{\sum_i(1-q_i)u_i}.
\]
It remains to show $\sum_i q_iu_i\cdot\sum_j(1-q_j)\le\sum_i q_i\cdot\sum_j(1-q_j)u_j$. The difference of the two sides is
\[
\sum_{i,j}q_i(1-q_j)(u_i-u_j)=\tfrac12\sum_{i,j}(u_i-u_j)(q_i-q_j)\le 0,
\]
where the equality follows by adding the expression to itself with $i$ and $j$ exchanged, and the inequality holds because $u_i=1/(1+\alpha q_i)$ is decreasing in $q_i$. Therefore $\mathrm{CGAR}(\mathcal A)\le(1+\alpha)\,\mathrm{CGAR}(\tilde{\mathcal A})$, with equality when all $q_i$ are equal.
 
\textbf{(ii) Correction from above.} The within-group entries are multiplied by $1+\alpha$, so $Z_i=1+\alpha(1-q_i)$ and $q_i'=q_i/(1+\alpha(1-q_i))<q_i$. Exchanging the roles of $q_i$ and $1-q_i$ in (i) gives $1/\mathrm{CGAR}(\tilde{\mathcal A})<1/\mathrm{CGAR}(\mathcal A)\le(1+\alpha)/\mathrm{CGAR}(\tilde{\mathcal A})$, which is the claim.
 
\textbf{(iii) No overshoot.} Let $x=\mathrm{nCGAR}(\tilde{\mathcal A})$ and $y=\mathrm{nCGAR}(\mathcal A)$. The condition $\alpha\le2\epsilon/(1-\epsilon)$ is equivalent to both $(1+\alpha)(1-\epsilon)\le1+\epsilon$ and $(1+\epsilon)/(1+\alpha)\ge1-\epsilon$. If $x<1-\epsilon$, then by (i) $x<y<(1+\alpha)(1-\epsilon)\le1+\epsilon$: $y$ moves toward the band and does not leave it on the other side, so its distance to the band, $\max(0,1-\epsilon-y)$, is strictly smaller than $1-\epsilon-x$. The case $x>1+\epsilon$ is symmetric by (ii). If $x\in[1-\epsilon,1+\epsilon]$, the target equals $\tilde{\mathcal A}$.
\end{proof}
 
\begin{remark}[Repeated correction]
Applying the correction $t$ times multiplies the cross-group entries by $(1+\alpha)^t$ before a single renormalization, so each $q_i\to1$ and $\mathrm{nCGAR}\to\infty$ as $t\to\infty$ (and $\mathrm{nCGAR}\to0$ in the symmetric case). Starting outside the band, $\mathrm{nCGAR}$ therefore reaches the band after finitely many steps, and by (iii) it does not jump over it. This is the fixed point used in the toy example (\autoref{fig:motivation}). FairPGT applies the correction once per training step.
\end{remark}
 
\subsubsection{Proof of \autoref{prop:align} (Attention Alignment)}
 
\begin{proof}
\textbf{(i)} Each term of $\mathcal L_{\mathrm{att}}$ is a KL divergence and is therefore nonnegative, and it is zero if and only if $\mathcal A_i=\tilde{\mathcal A}_i$. Consider a layer, head, and partition with at least two groups. If its $\mathrm{nCGAR}$ lies in the band, the target equals the attention and all its rows contribute zero. Otherwise, by \autoref{thm:cgar}(i)--(ii), $q_i'\neq q_i$ for every row with $q_i<1$, and at least one such row exists; hence $\mathcal A_i\neq\tilde{\mathcal A}_i$ for that row, and it contributes a positive term. Partitions with a single group are not in $\mathcal I$.
 
\textbf{(ii)} Write $\sigma=\mathrm{softmax}(a_i)=\tilde{\mathcal A}_i$ and $t=\mathcal A_i$. Since $t$ is constant and sums to one,
\[
\ell_i=\sum_j t_j\log t_j-\sum_j t_j a_j+\log\sum_k e^{a_k},\qquad
\nabla_{a_i}\ell_i=\sigma-t .
\]
The gradient step changes the logits by $\eta(t-\sigma)$. The cross-group mass $q_i=\sum_{j\in c(i)}\sigma_j$ has partial derivatives $\partial q_i/\partial a_k=\sigma_k(\mathbb {I}[k\in c(i)]-q_i)$, so
\[
\frac{dq_i}{d\eta}\Big|_{\eta=0}=(1-q_i)\sum_{j\in c(i)}\sigma_j(t_j-\sigma_j)\;-\;q_i\sum_{j\in w(i)}\sigma_j(t_j-\sigma_j).
\]
When cross-group entries are boosted, the target of \autoref{thm:cgar}(i) gives $t_j-\sigma_j=\sigma_j\alpha(1-q_i)/(1+\alpha q_i)$ for $j\in c(i)$ and $t_j-\sigma_j=-\sigma_j\alpha q_i/(1+\alpha q_i)$ for $j\in w(i)$. Substituting gives the stated rate, which is positive since $q_i\in(0,1)$ and both sums contain positive terms. When within-group entries are boosted, $t_j-\sigma_j=-\sigma_j\alpha(1-q_i)/(1+\alpha(1-q_i))$ for $j\in c(i)$ and $t_j-\sigma_j=\sigma_j\alpha q_i/(1+\alpha(1-q_i))$ for $j\in w(i)$, which gives the same bracket with factor $-\alpha/(1+\alpha(1-q_i))$.
\end{proof}
 
\subsubsection{Proof of \autoref{thm:target} (Graph-Aware Target)}
 
\begin{proof}
Let $\delta_{ik}=M_{ik}-\bar M_i$, so that $\tilde M_{ik}=\tanh(\delta_{ik})\in(-1,1)$ and $e^{\tilde M_{ik}}\in(e^{-1},e)$. Student-$t$ assignments are positive, so $P^{\mathrm{DEC}}_{ik}>0$. Dividing the numerator and denominator of Eq.~\eqref{eq:unified-target} by $\sum_{k'}C_{ik'}^2/f_{k'}$ gives
\begin{equation}
P_{ik}=\frac{P^{\mathrm{DEC}}_{ik}\,e^{\tilde M_{ik}}}{\sum_{k'}P^{\mathrm{DEC}}_{ik'}\,e^{\tilde M_{ik'}}}.
\label{eq:target-dec}
\end{equation}
 
\textbf{(i)} The denominator of Eq.~\eqref{eq:target-dec} is a convex combination of values in $(e^{-1},e)$ and therefore lies in $(e^{-1},e)$. Hence $P_{ik}/P^{\mathrm{DEC}}_{ik}=e^{\tilde M_{ik}}/\sum_{k'}P^{\mathrm{DEC}}_{ik'}e^{\tilde M_{ik'}}\in(e^{-2},e^{2})$.
 
\textbf{(ii)} The ratio follows from Eq.~\eqref{eq:target-dec}, since the denominator does not depend on $k$. Centering subtracts the same $\bar M_i$ from $M_{ik}$ and $M_{ik'}$, and $\tanh$ is strictly increasing, so $\tilde M_{ik}-\tilde M_{ik'}$ has the sign of $M_{ik}-M_{ik'}$ and lies in $(-2,2)$. If $M_{ik}>M_{ik'}$ and $P^{\mathrm{DEC}}_{ik}\ge P^{\mathrm{DEC}}_{ik'}$, both factors of the ratio are at least $1$ and the second is larger than $1$, so $P_{ik}>P_{ik'}$. If $P^{\mathrm{DEC}}_{ik}\ge P^{\mathrm{DEC}}_{ik'}$ but $P_{ik'}>P_{ik}$, then $P^{\mathrm{DEC}}_{ik}/P^{\mathrm{DEC}}_{ik'}<\exp(\tilde M_{ik'}-\tilde M_{ik})<e^2$, and the exponent is positive, so $M_{ik'}>M_{ik}$.
 
\textbf{(iii)} If $M_{ik}$ is the same for all $k$, then $\delta_{ik}=0$ and $e^{\tilde M_{ik}}=1$ for all $k$, and Eq.~\eqref{eq:target-dec} reduces to $P^{\mathrm{DEC}}_{ik}$.
\end{proof}
 
\begin{remark}[Centering]
For row-stochastic $C$, $\sum_k M_{ik}=\sum_j A_{ij}\sum_k C_{jk}-\frac{d_i}{2|\mathcal E|}\sum_j d_j\sum_k C_{jk}=d_i-d_i=0$, using $\sum_j d_j=2|\mathcal E|$. Hence $\bar M_i=0$ and $\tilde M_{ik}=\tanh(M_{ik})$; the explicit centering only guards against numerical drift. In the case of (iii), all gains of node $i$ are therefore zero.
\end{remark}

\subsection{Proof of \autoref{thm:complexity} (Complexity of FairPGT)}
 
We count multiply--add operations of the forward pass; the backward pass has the same order. Partitions are balanced, $n_p=n/N_p$, and $D=Hd_h$.

\begin{proof}
\textbf{(i) One encoder layer.} \emph{Projections and feed-forward network:} the query, key, value, and output projections and the two-layer feed-forward network with hidden width $\mathcal{O}(D)$ cost $\mathcal{O}(nD^2)$. \emph{Local attention:} in each partition and head, $Q_pK_p^\top$ and the product with $V_p$ cost $\mathcal{O}(n_p^2d_h)$; over $H$ heads and $N_p$ partitions, this gives $N_pH\,\mathcal{O}(n^2d_h/N_p^2)=\mathcal{O}(n^2D/N_p)$. \emph{Routing tokens:} mean pooling into $N_p$ tokens costs $\mathcal{O}(nD)$; self-attention among the tokens costs $\mathcal{O}(N_p^2D)$, plus $\mathcal{O}(N_pD^2)$ for their projections, which is dominated by $\mathcal{O}(nD^2)$; and the broadcast, in which each node attends to all $N_p$ tokens, costs $\mathcal{O}(nN_pD)$. \emph{Gate, residuals, and normalization:} these cost $\mathcal{O}(nD)$. Summing all terms gives (i).

\textbf{(ii) Training step.} The input embedding costs $\mathcal{O}(nd_0D)$, and the $L$ layers cost $L$ times (i). The projection $z_i=\mathrm{LN}(W_zh_i)$ costs $\mathcal{O}(nDd_z)$, and the DEC assignments cost $\mathcal{O}(nKd_z)$. In Eq.~\eqref{eq:mod-gain}, the neighbor term $\sum_{j\in\mathcal N(i)}C_{jk}$ for all $i,k$ is a sparse--dense product costing $\mathcal{O}(|\mathcal E|K)$, and the null-model term needs the $K$ sums $\sum_jd_jC_{jk}$ only once, so it costs $\mathcal{O}(nK)$; centering, $\tanh$, and the normalization in Eq.~\eqref{eq:unified-target} also cost $\mathcal{O}(nK)$. The losses $\mathcal{L}_{\mathrm{clust}}$, $\mathcal{L}_{\mathrm{grp}}$, and $\mathcal{L}_{\mathrm{info}}$ each cost $\mathcal{O}(nK)$; for $\mathcal L_{\mathrm{grp}}$, each node adds its row of $C$ to the column of its group. The CGAR target and $\mathcal{L}_{\mathrm{att}}$ need two passes over each partition attention matrix, $\mathcal{O}(Hn^2/N_p)$ per layer, which is dominated by local attention since $H\le D$. For memory, back-propagation stores the partition attention matrices, $\mathcal{O}(Hn^2/N_p)$ per layer, the broadcast attention, $\mathcal{O}(HnN_p)$ per layer, and the node activations, $\mathcal{O}(nD)$ per layer, while the graph and the matrices $C$ and $P$ need $\mathcal{O}(|\mathcal E|+nK)$.

\textbf{(iii) Inference.} The encoder, projection, and assignments are computed as in (ii), but the targets and the losses are not. Without back-propagation, the attention matrices of a layer can be released before the next layer is computed, so memory is $\mathcal{O}(Hn^2/N_p+nD)$.

\textbf{Choice of $N_p$.} Balancing the local-attention term $n^2D/N_p$ against the broadcast term $nN_pD$ gives $N_p=\sqrt n$, for which both are $n^{3/2}D$ and the token self-attention is $nD$. One layer then costs $\mathcal{O}(n^{3/2}D+nD^2)$, compared to $\mathcal{O}(n^2D+nD^2)$ for global attention.
\end{proof}
 
\paragraph{Example}
On Pokec-n ($n=66{,}569$), our default gives $N_p=27$ partitions of about $2{,}470$ nodes, so the partition attention matrices of one head in one layer hold about $27\cdot2{,}470^2\approx1.6\times10^{8}$ entries, compared to $n^2\approx4.4\times10^{9}$ for global attention, a $27\times$ reduction.

\section{Experiments}\label{sec:experiments}

Our experiments are designed to test the two main claims of this paper: that a graph transformer can cluster fairly without giving up the structure that is unrelated to the protected groups, and that its attention can be kept group-neutral at the same time. To this end, we organize them around four questions.
\textbf{Q1}: How does FairPGT trade clustering quality against group fairness, compared with fairness-agnostic and fairness-aware baselines?
\textbf{Q2}: How does the price of fairness depend on the alignment between communities and protected groups, and does the encoder stay group-neutral?
\textbf{Q3}: Which components are responsible for the fairness, and which for the clustering quality (ablation)? 
\textbf{Q4}: What do partitioning and the fairness weight cost, and what do they buy (sensitivity and cost)?

\subsection{Setup}\label{sec:setup}

\paragraph{\textbf{Datasets}}
We use seven attributed graphs that span social, financial, and music-listening networks and range from 391 to 67{,}796 nodes (\autoref{tab:datasets}). Throughout, class labels are used only for evaluation, and the protected attribute is removed from the node features of every dataset, so that all methods are exposed to the same input; the only exception is FairGT, whose design requires it (see Baselines).
NBA~\citep{DBLP:conf/wsdm/DaiW21} connects basketball players who are linked on Twitter; the protected attribute is nationality (US or not), and the label is the salary class used by Dai and Wang~\citep{DBLP:conf/wsdm/DaiW21}, binarized as majority class versus the rest.
Pokec-n and Pokec-z~\citep{takac2012data,DBLP:conf/wsdm/DaiW21} are two regional subgraphs of the Slovak social network Pokec. Here, the protected attribute is age in four groups ($\le 18$, 19--25, 26--35, $>35$), so these two graphs test fairness beyond two groups; the label is the working field, binarized as majority field versus the rest. Profile attributes are one-hot encoded and reduced to 128 dimensions by truncated SVD, and isolated nodes are removed.
German and Credit~\citep{DBLP:journals/widm/QuyRIZN22} are credit-scoring datasets, with gender and marital status as protected attributes and creditworthiness and default as labels. For Credit, we use the customer similarity graph of Agarwal et al.~\citep{DBLP:conf/uai/AgarwalLZ21}; for German, we build a symmetric 10-nearest-neighbour graph with cosine similarity on the non-protected features (gender and age excluded). In both cases, edges therefore reflect feature similarity rather than observed interactions.

LastFM~\citep{DBLP:conf/cikm/RozemberczkiS20} links music-platform users who follow each other; its features are the artists a user likes, reduced to 128 dimensions by truncated SVD, and its labels are the users' countries, restricted to the six largest. Similarly, Facebook~\citep{DBLP:journals/compnet/RozemberczkiAS21} links verified Facebook pages that like each other, with the page category (four classes) as label and keyword features from the page descriptions.
Since neither graph has a demographic attribute, we construct a binary group by $k$-means ($k=2$) on the node features. This makes the group a function of the features, which is the strongest proxy setting an encoder can face; moreover, edges also follow these groups ($h_S=0.73$ and $0.64$ in \autoref{tab:datasets}, against about $0.5$ for groups unrelated to the edges).

Finally, for the controlled study of \autoref{sec:sbm}, we generate attributed stochastic block models (SBMs)~\citep{DBLP:journals/jmlr/Abbe17} with node features in the style of contextual SBMs~\citep{DBLP:conf/nips/DeshpandeSMM18}, and with planted protected groups as in the block models used to analyse fair spectral clustering~\citep{kleindessner2019fair,DBLP:conf/nips/GuptaD22}. Each graph has $n=2000$ nodes, $K=4$ equal communities, and two groups. A node in community $k$ belongs to group $k \bmod 2$ with probability $a$ and otherwise follows the population share (minority 0.3), so the alignment $a$ runs from 0, where groups are independent of the communities, to 1, where every community is one group. Edges are eight times more likely within than across communities (mean degree 12), and the 32-dimensional features combine a community signal, a group proxy of equal strength, and Gaussian noise.

\begin{table*}[t]
\centering
\footnotesize
\caption{Dataset statistics. $N$: nodes; $|E|$: undirected edges; $d$: feature dimension; $K$: label classes (number of clusters); $G$: protected groups; Maj.: share of the largest group; $h_S$, $h_Y$: fraction of edges joining nodes of the same group and of the same class; CC: connected components.}
\label{tab:datasets}
\adjustbox{max width=\textwidth}{
\begin{tabular}{lrrrcccccrll}
\toprule
Dataset & $N$ & $|E|$ & $d$ & $K$ & $G$ & Maj. & $h_S$ & $h_Y$ & CC & Protected attribute & Label \\
\midrule
NBA & 391 & 10,416 & 95 & 2 & 2 & 73.7\% & 0.73 & 0.51 & 1 & Nationality (US / non-US) & Salary class (majority vs.\ rest) \\
German & 1,000 & 7,156 & 39 & 2 & 2 & 69.0\% & 0.78 & 0.64 & 1 & Gender & Good / bad credit \\
LastFM & 5,465 & 19,024 & 128 & 6 & 2 & 53.1\% & 0.73 & 0.92 & 16 & Synthetic (feature $k$-means) & Country (6) \\
Facebook & 22,470 & 170,823 & 4,714 & 4 & 2 & 60.1\% & 0.64 & 0.89 & 1 & Synthetic (feature $k$-means) & Page type (4) \\
Credit & 30,000 & 137,377 & 12 & 2 & 2 & 54.5\% & 0.65 & 0.64 & 120 & Marital status & Default next month \\
Pokec-n & 66,569 & 517,047 & 128 & 2 & 4 & 46.2\% & 0.58 & 0.75 & 226 & Age (4 bins) & Working field (majority vs.\ rest) \\
Pokec-z & 67,796 & 617,958 & 128 & 2 & 4 & 47.1\% & 0.61 & 0.73 & 170 & Age (4 bins) & Working field (majority vs.\ rest) \\
\midrule
SBM ($a\in[0,1]$) & 2,000 & 11,925 & 32 & 4 & 2 & 50.0--69.7\% & 0.56--0.82 & 0.73 & 1 & Planted, aligned by $a$ & Planted community \\
\bottomrule
\end{tabular}}
\end{table*}

\paragraph{\textbf{Baselines}}
The baselines cover the method families reviewed in \autoref{sec:related} (\autoref{tab:comparison}). As fairness-agnostic deep graph clustering methods, we use DMoN~\citep{DBLP:journals/jmlr/TsitsulinPPM23}, SDCN~\citep{DBLP:conf/www/Bo0SZL020}, CCGC~\citep{DBLP:conf/aaai/YangLZWTZLFZ23}, SCGC~\citep{DBLP:journals/tnn/LiuYZLWLTL24}, DinkNet~\citep{DBLP:conf/icml/dinknet23}, DiffPool~\citep{DBLP:conf/nips/DiffPool18}, and MinCutPool~\citep{DBLP:conf/icml/BianchiGA20}. As fair deep models, we use DMoN-SP, which adds a statistical-parity penalty to DMoN in the spirit of modularity-fair community detection~\citep{DBLP:conf/icdm/GkartziosPT25}; FairGC~\citep{DBLP:conf/aaai/ZhangWRWYMMLJ26}; and the fair graph transformers FairGT~\citep{DBLP:conf/ijcai/LuoHY0024} and FairGP~\citep{DBLP:conf/aaai/LuoH0X0025}. Finally, as fair spectral clustering, we use sFairSC, group-fair spectral clustering~\citep{kleindessner2019fair} in its scalable form~\citep{wang2023scalable}, and iFairSC, individual-fair spectral clustering~\citep{DBLP:conf/nips/GuptaD22}.

Some of these baselines needed adaptation, most of all FairGT and FairGP. Both were proposed for node classification and obtain fairness through their architectures rather than through a loss: FairGT feeds the protected attribute into its encodings, and FairGP removes attention across METIS partitions. We re-implement these mechanisms (for FairGT, with neighbourhood attention layers instead of hop-token sequences), attach a clustering head trained by DEC-style self-training~\citep{xie2016dec}, and add the statistical-parity penalty of DMoN-SP, so that both act as fairness-aware clustering methods; FairGT receives the protected attribute in training and at inference, as in the original. For CCGC, we use GCN encoders in place of its filtered-feature MLPs.
For the spectral methods, sFairSC is computed on the regularized graph $A+(\tau/n)\mathbf{1}\mathbf{1}^\top$~\citep{DBLP:conf/nips/QinR13}, $\tau$ the mean degree, because the leading eigenvectors of graphs with many connected components (LastFM, Credit, Pokec) are otherwise component indicators, and $k$-means then returns a single cluster. iFairSC uses a symmetric 10-nearest-neighbour representation graph on the features, with the low-rank approximation of the original paper when the constraint is infeasible. Lastly, DMoN-SP uses a penalty weight of $0.1$; at this weight it behaves almost like DMoN (\autoref{tab:main}), and we report it unchanged rather than tune it on labels.
We do not compare with fairlets~\citep{chierichetti2017fair}, which address $k$-center and $k$-median clustering of points without a graph, or with fair graph neural networks for node classification~\citep{DBLP:conf/wsdm/DaiW21,DBLP:conf/uai/AgarwalLZ21,DBLP:conf/www/DongLJL22}, which need labels for training.

\paragraph{\textbf{Protocol}}
Every method runs for 200 epochs with one fixed hyperparameter setting across all datasets, and no hyperparameter is tuned on labels. Likewise, we report the clustering at the last epoch, so that no label information is used for model selection. $K$ equals the number of label classes, and results are the mean and standard deviation over five seeds (three on the SBMs).
Methods whose memory grows quadratically or cubically with $n$ are skipped above a size limit (DiffPool and MinCutPool above $5\times10^4$ nodes, FairGC above $2.5\times10^4$, iFairSC above $8\times10^3$) and marked \textit{n/s}, while runs that exceed the GPU memory are marked \textit{OOM}.
Finally, we call a run \emph{collapsed} if it uses fewer than $K$ clusters or puts more than 90\% of the nodes into one cluster. Collapsed results are marked and not ranked on the real graphs, because a single cluster is trivially fair.

\paragraph{\textbf{Metrics}}
For clustering quality, we report normalized mutual information (NMI)~\citep{DBLP:journals/jmlr/StrehlG02} and macro-F1~\citep{DBLP:journals/ipm/SokolovaL09} between clusters and labels, with clusters matched to classes by the Hungarian algorithm~\citep{DBLP:books/daglib/p/Kuhn10} for F1, as well as modularity $Q$~\citep{newman2006modularity}, which measures community structure without labels. For fairness, we report the statistical-parity gap~\citep{DBLP:conf/innovations/DworkHPRZ12} of the hard clustering,
\begin{equation}
\Delta_{\mathrm{SP}}=\max_{k}\max_{g}\big|\Pr(s=g\mid c=k)-\Pr(s=g)\big|,
\end{equation}
that is, the largest deviation of any cluster's group share from the population share; it is zero exactly when the clusters are independent of the protected attribute, for any number of groups. Higher is better ($\uparrow$) for NMI, F1, and $Q$, and lower is better ($\downarrow$) for $\Delta_{\mathrm{SP}}$.
Beyond the clusters, two further quantities describe the trained encoder itself. The first is the nCGAR of the trained attention (\autoref{def:cgar}), averaged over layers, heads, and partitions with at least two groups, where 1 means group-neutral attention. The second, which we call the \emph{$S$-probe}, measures how much of the protected attribute remains in the node embeddings: a logistic-regression probe~\citep{DBLP:conf/iclr/AlainB17} is trained to predict $s_i$ from $z_i$, as done to measure protected information in graph embeddings~\citep{DBLP:conf/icml/BoseH19}, and we report its balanced accuracy~\citep{DBLP:conf/icpr/BrodersenOSB10} under 5-fold cross-validation. We use balanced accuracy because the groups are imbalanced; a value of 1/G (0.5 for two groups) means that $S$ cannot be linearly decoded ($\downarrow$).

\paragraph{\textbf{Implementation}}
FairPGT uses three layers with hidden size 256 and four heads, eight spectral encodings, a 32-dimensional clustering projection, dropout 0.1, and $N_p=\max(8,\lceil n/2500\rceil)$ METIS partitions. Input features are standardized, and cluster centres are initialized by $k$-means on the initial embeddings. The CGAR target uses $\alpha=\epsilon=0.05$, the graph-aware target is refreshed every $T_P=5$ epochs, and the loss weights are $\lambda_{\mathrm{grp}}=10$, $\lambda_{\mathrm{att}}=1$, and $\lambda_{\mathrm{info}}=1$. We train with Adam~\citep{kingma2014adam} (learning rate $5\times10^{-4}$, weight decay $10^{-5}$, gradient clipping at 1). Baselines use the default hyperparameters of their reference implementations where available, aligned only in the number of clusters, the epoch budget, and the evaluation protocol. All methods are implemented in PyTorch and PyTorch Geometric, and each run uses a single NVIDIA GPU (H100 NVL for the largest graphs). %

\subsection{Q1: Quality and fairness on real graphs}\label{sec:main-results}

\begin{figure*}[t]
  \centering
  \includegraphics[width=\textwidth]{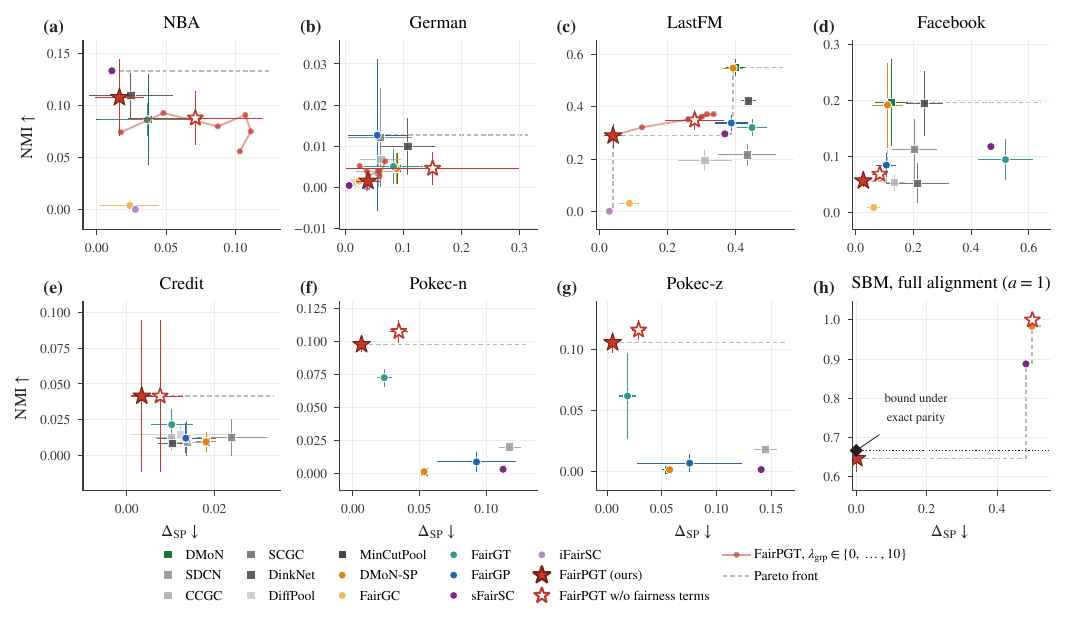}
  \caption{Quality--fairness plane (mean $\pm$ std over five seeds; up and left is better). Squares: fairness-agnostic baselines; circles: fair baselines; filled star: FairPGT; hollow star: FairPGT without fairness terms. The dashed line is the Pareto front of the methods, and the light red line traces FairPGT for $\lambda_{\mathrm{grp}}\in\{0,0.1,0.3,1,3,10\}$ on NBA, German and LastFM. (h)~SBM at full alignment; the diamond is the best NMI any clustering with $\Delta_{\mathrm{SP}}=0$ can reach (Proposition~\ref{prop:price}). Collapsed runs on the real graphs are omitted.}
  \label{fig:tradeoff}
\end{figure*}

\autoref{tab:main} reports NMI and $\Delta_{\mathrm{SP}}$ for all methods, and \autoref{tab:extra} complements it with macro-F1 and modularity. To make the trade-off visible at a glance, \autoref{fig:tradeoff} places every method in the NMI--$\Delta_{\mathrm{SP}}$ plane, together with FairPGT trained without its fairness terms.

\begin{table*}[t]
\centering
\caption{Clustering quality (NMI) and group fairness ($\Delta_{\mathrm{SP}}$) on seven benchmarks. Mean and standard deviation over five seeds. Best in \textbf{\underline{bold and underlined}}, runner-up in \underline{underlined}, among runs that did not collapse; $^\dagger$: collapsed (fewer than $K$ clusters used, or one cluster holding over 90\% of the nodes, in at least half of the runs). \textit{n/s}: not scalable, skipped (memory quadratic or cubic in $n$); \textit{OOM}: out of GPU memory.}
\label{tab:main}
\adjustbox{max width=\textwidth}{
\setlength{\tabcolsep}{2.6pt}
\footnotesize
\begin{tabular}{ll|ccccccc|cccc|cc|c}
\toprule
& & \multicolumn{7}{c}{\textit{Deep graph clustering}} & \multicolumn{4}{c}{\textit{Fair deep}} & \multicolumn{2}{c}{\textit{Fair spectral}} & \textit{Ours} \\
\cmidrule(lr){3-9} \cmidrule(lr){10-13} \cmidrule(lr){14-15} \cmidrule(lr){16-16}
& & DMoN & SDCN & CCGC & SCGC & DinkNet & DiffPool & MinCutPool & DMoN-SP & FairGC & FairGT & FairGP & sFairSC & iFairSC & \textbf{FairPGT} \\
\midrule
\multirow{2}{*}{\textbf{NBA}} & NMI$\uparrow$ & .086\std{.04} & .000\std{.00}$^\dagger$ & .036\std{.05}$^\dagger$ & .021\std{.03}$^\dagger$ & .000\std{.00}$^\dagger$ & .014\std{.03}$^\dagger$ & \underline{.109}\std{.02} & .047\std{.05}$^\dagger$ & .004\std{.01} & .086\std{.02} & .016\std{.01}$^\dagger$ & \textbf{\underline{.133}}\std{.00} & .000\std{.00} & .107\std{.04} \\
 & $\Delta_{\mathrm{SP}}$$\downarrow$ & .037\std{.04} & .000\std{.00}$^\dagger$ & .116\std{.10}$^\dagger$ & .255\std{.29}$^\dagger$ & .000\std{.00}$^\dagger$ & .058\std{.12}$^\dagger$ & .025\std{.03} & .079\std{.12}$^\dagger$ & .024\std{.02} & .036\std{.03} & .125\std{.05}$^\dagger$ & \textbf{\underline{.011}}\std{.00} & .028\std{.00} & \underline{.016}\std{.02} \\
\midrule
\multirow{2}{*}{\textbf{German}} & NMI$\uparrow$ & .005\std{.00} & .004\std{.00} & .007\std{.01} & \underline{.012}\std{.01} & .000\std{.00}$^\dagger$ & .001\std{.00} & .010\std{.01} & .005\std{.00} & .001\std{.00} & .005\std{.00} & \textbf{\underline{.013}}\std{.02} & .000\std{.00} & .000\std{.00} & .002\std{.00} \\
 & $\Delta_{\mathrm{SP}}$$\downarrow$ & .090\std{.04} & .057\std{.04} & .061\std{.04} & .060\std{.06} & .000\std{.00}$^\dagger$ & \underline{.016}\std{.02} & .108\std{.05} & .088\std{.04} & .023\std{.02} & .083\std{.05} & .055\std{.05} & \textbf{\underline{.006}}\std{.00} & .038\std{.00} & .039\std{.02} \\
\midrule
\multirow{2}{*}{\textbf{LastFM}} & NMI$\uparrow$ & \textbf{\underline{.550}}\std{.03} & .405\std{.09}$^\dagger$ & .194\std{.04} & .216\std{.04} & .081\std{.12}$^\dagger$ & .207\std{.08}$^\dagger$ & .423\std{.02} & \underline{.547}\std{.02} & .032\std{.01} & .321\std{.03} & .338\std{.03} & .296\std{.00} & .001\std{.00} & .289\std{.05} \\
 & $\Delta_{\mathrm{SP}}$$\downarrow$ & .399\std{.03} & .218\std{.10}$^\dagger$ & .309\std{.08} & .433\std{.09} & .515\std{.03}$^\dagger$ & .281\std{.04}$^\dagger$ & .436\std{.02} & .392\std{.03} & .088\std{.03} & .448\std{.04} & .387\std{.05} & .368\std{.00} & \textbf{\underline{.029}}\std{.01} & \underline{.040}\std{.03} \\
\midrule
\multirow{2}{*}{\textbf{Facebook}} & NMI$\uparrow$ & \textbf{\underline{.196}}\std{.08} & .080\std{.18}$^\dagger$ & .054\std{.01} & .113\std{.05} & .053\std{.04} & .068\std{.04}$^\dagger$ & \underline{.195}\std{.06} & .192\std{.08} & .010\std{.01} & .095\std{.04} & .085\std{.02} & .118\std{.00} & \textit{n/s} & .057\std{.01} \\
 & $\Delta_{\mathrm{SP}}$$\downarrow$ & .124\std{.06} & .025\std{.06}$^\dagger$ & .133\std{.04} & .204\std{.08} & .214\std{.11} & .329\std{.12}$^\dagger$ & .238\std{.06} & .110\std{.05} & \underline{.063}\std{.02} & .519\std{.09} & .106\std{.03} & .468\std{.00} & \textit{n/s} & \textbf{\underline{.027}}\std{.01} \\
\midrule
\multirow{2}{*}{\textbf{Credit}} & NMI$\uparrow$ & .010\std{.01} & .009\std{.00} & .013\std{.01} & .013\std{.01} & .012\std{.01} & .015\std{.01} & .008\std{.00} & .010\std{.01} & \textit{n/s} & \underline{.022}\std{.01} & .012\std{.01} & .002\std{.00}$^\dagger$ & \textit{n/s} & \textbf{\underline{.042}}\std{.05} \\
 & $\Delta_{\mathrm{SP}}$$\downarrow$ & .018\std{.00} & .014\std{.00} & \underline{.010}\std{.00} & .024\std{.01} & .014\std{.01} & .012\std{.01} & .010\std{.00} & .018\std{.00} & \textit{n/s} & .010\std{.00} & .013\std{.01} & .030\std{.00}$^\dagger$ & \textit{n/s} & \textbf{\underline{.003}}\std{.00} \\
\midrule
\multirow{2}{*}{\textbf{Pokec-n}} & NMI$\uparrow$ & .001\std{.00} & .020\std{.00} & \textit{OOM} & \textit{OOM} & .011\std{.00}$^\dagger$ & \textit{n/s} & \textit{n/s} & .001\std{.00} & \textit{n/s} & \underline{.072}\std{.01} & .009\std{.01} & .003\std{.00} & \textit{n/s} & \textbf{\underline{.098}}\std{.01} \\
 & $\Delta_{\mathrm{SP}}$$\downarrow$ & .054\std{.00} & .118\std{.01} & \textit{OOM} & \textit{OOM} & .102\std{.03}$^\dagger$ & \textit{n/s} & \textit{n/s} & .053\std{.00} & \textit{n/s} & \underline{.023}\std{.01} & .093\std{.03} & .113\std{.00} & \textit{n/s} & \textbf{\underline{.006}}\std{.00} \\
\midrule
\multirow{2}{*}{\textbf{Pokec-z}} & NMI$\uparrow$ & .001\std{.00} & .018\std{.00} & \textit{OOM} & \textit{OOM} & .015\std{.00}$^\dagger$ & \textit{n/s} & \textit{n/s} & .001\std{.00} & \textit{n/s} & \underline{.062}\std{.04} & .007\std{.01} & .001\std{.00} & \textit{n/s} & \textbf{\underline{.106}}\std{.01} \\
 & $\Delta_{\mathrm{SP}}$$\downarrow$ & .056\std{.01} & .145\std{.01} & \textit{OOM} & \textit{OOM} & .143\std{.02}$^\dagger$ & \textit{n/s} & \textit{n/s} & .057\std{.00} & \textit{n/s} & \underline{.019}\std{.01} & .075\std{.05} & .141\std{.00} & \textit{n/s} & \textbf{\underline{.005}}\std{.00} \\
\bottomrule
\end{tabular}}
\end{table*}

\begin{table*}[t]
\centering
\caption{Complementary quality metrics: macro-F1 against the labels and modularity $Q$ of the graph. Mean and standard deviation over five seeds. Best in \textbf{\underline{bold and underlined}}, runner-up in \underline{underlined}, among runs that did not collapse; $^\dagger$: collapsed (fewer than $K$ clusters used, or one cluster holding over 90\% of the nodes, in at least half of the runs). \textit{n/s}: not scalable, skipped (memory quadratic or cubic in $n$); \textit{OOM}: out of GPU memory.}
\label{tab:extra}
\adjustbox{max width=\textwidth}{
\setlength{\tabcolsep}{2.6pt}
\footnotesize
\begin{tabular}{ll|ccccccc|cccc|cc|c}
\toprule
& & \multicolumn{7}{c}{\textit{Deep graph clustering}} & \multicolumn{4}{c}{\textit{Fair deep}} & \multicolumn{2}{c}{\textit{Fair spectral}} & \textit{Ours} \\
\cmidrule(lr){3-9} \cmidrule(lr){10-13} \cmidrule(lr){14-15} \cmidrule(lr){16-16}
& & DMoN & SDCN & CCGC & SCGC & DinkNet & DiffPool & MinCutPool & DMoN-SP & FairGC & FairGT & FairGP & sFairSC & iFairSC & \textbf{FairPGT} \\
\midrule
\multirow{2}{*}{\textbf{NBA}} & F1$\uparrow$ & .626\std{.04} & .376\std{.00}$^\dagger$ & .476\std{.13}$^\dagger$ & .372\std{.01}$^\dagger$ & .376\std{.00}$^\dagger$ & .371\std{.01}$^\dagger$ & .593\std{.08} & .488\std{.16}$^\dagger$ & .520\std{.03} & .659\std{.02} & .459\std{.10}$^\dagger$ & \textbf{\underline{.690}}\std{.00} & .497\std{.00} & \underline{.677}\std{.03} \\
 & $Q$$\uparrow$ & \underline{.105}\std{.01} & .000\std{.00}$^\dagger$ & .025\std{.04}$^\dagger$ & .001\std{.00}$^\dagger$ & .000\std{.00}$^\dagger$ & .001\std{.00}$^\dagger$ & .072\std{.04} & .040\std{.05}$^\dagger$ & .023\std{.01} & .005\std{.00} & .023\std{.02}$^\dagger$ & \textbf{\underline{.127}}\std{.00} & $-$.003\std{.00} & .070\std{.02} \\
\midrule
\multirow{2}{*}{\textbf{German}} & F1$\uparrow$ & .524\std{.01} & .489\std{.05} & .524\std{.02} & .520\std{.04} & .412\std{.00}$^\dagger$ & .472\std{.03} & \textbf{\underline{.541}}\std{.02} & .524\std{.01} & .494\std{.01} & .519\std{.03} & \underline{.527}\std{.05} & .488\std{.00} & .484\std{.00} & .491\std{.02} \\
 & $Q$$\uparrow$ & \underline{.421}\std{.01} & .237\std{.16} & .364\std{.01} & .330\std{.02} & .000\std{.00}$^\dagger$ & .228\std{.15} & .394\std{.01} & \textbf{\underline{.422}}\std{.01} & .214\std{.05} & .326\std{.02} & .353\std{.01} & .391\std{.00} & $-$.010\std{.00} & .418\std{.03} \\
\midrule
\multirow{2}{*}{\textbf{LastFM}} & F1$\uparrow$ & \textbf{\underline{.589}}\std{.05} & .318\std{.07}$^\dagger$ & .352\std{.03} & .335\std{.06} & .133\std{.09}$^\dagger$ & .188\std{.05}$^\dagger$ & .456\std{.04} & \underline{.588}\std{.04} & .241\std{.01} & .457\std{.05} & .443\std{.04} & .319\std{.00} & .171\std{.00} & .403\std{.04} \\
 & $Q$$\uparrow$ & \textbf{\underline{.666}}\std{.03} & .542\std{.08}$^\dagger$ & .487\std{.03} & .405\std{.06} & .113\std{.15}$^\dagger$ & .262\std{.09}$^\dagger$ & .614\std{.02} & \underline{.661}\std{.02} & .139\std{.02} & .471\std{.03} & .609\std{.03} & .555\std{.00} & $-$.002\std{.00} & .460\std{.05} \\
\midrule
\multirow{2}{*}{\textbf{Facebook}} & F1$\uparrow$ & \textbf{\underline{.493}}\std{.08} & .196\std{.18}$^\dagger$ & .369\std{.01} & .416\std{.06} & .276\std{.06} & .274\std{.04}$^\dagger$ & .449\std{.06} & \underline{.493}\std{.08} & .288\std{.02} & .389\std{.04} & .375\std{.03} & .253\std{.00} & \textit{n/s} & .373\std{.02} \\
 & $Q$$\uparrow$ & \underline{.648}\std{.01} & .107\std{.24}$^\dagger$ & .556\std{.02} & .568\std{.03} & .376\std{.13} & .350\std{.12}$^\dagger$ & .606\std{.03} & \textbf{\underline{.648}}\std{.02} & .208\std{.06} & .396\std{.11} & .596\std{.02} & .409\std{.00} & \textit{n/s} & .592\std{.02} \\
\midrule
\multirow{2}{*}{\textbf{Credit}} & F1$\uparrow$ & .468\std{.05} & .477\std{.02} & .507\std{.02} & .514\std{.04} & .489\std{.05} & .497\std{.05} & .494\std{.01} & .468\std{.05} & \textit{n/s} & \underline{.527}\std{.02} & .505\std{.03} & .452\std{.00}$^\dagger$ & \textit{n/s} & \textbf{\underline{.536}}\std{.12} \\
 & $Q$$\uparrow$ & \underline{.462}\std{.00} & \textbf{\underline{.472}}\std{.01} & .414\std{.03} & .374\std{.02} & .412\std{.07} & .393\std{.04} & .443\std{.01} & .462\std{.00} & \textit{n/s} & .382\std{.04} & .440\std{.02} & .236\std{.00}$^\dagger$ & \textit{n/s} & .368\std{.09} \\
\midrule
\multirow{2}{*}{\textbf{Pokec-n}} & F1$\uparrow$ & .428\std{.01} & .445\std{.03} & \textit{OOM} & \textit{OOM} & .459\std{.00}$^\dagger$ & \textit{n/s} & \textit{n/s} & .431\std{.00} & \textit{n/s} & \textbf{\underline{.525}}\std{.00} & .448\std{.03} & \underline{.525}\std{.00} & \textit{n/s} & .453\std{.06} \\
 & $Q$$\uparrow$ & \underline{.411}\std{.00} & .186\std{.01} & \textit{OOM} & \textit{OOM} & .004\std{.00}$^\dagger$ & \textit{n/s} & \textit{n/s} & \textbf{\underline{.411}}\std{.00} & \textit{n/s} & .012\std{.01} & .266\std{.07} & .320\std{.00} & \textit{n/s} & .059\std{.03} \\
\midrule
\multirow{2}{*}{\textbf{Pokec-z}} & F1$\uparrow$ & .434\std{.01} & .443\std{.03} & \textit{OOM} & \textit{OOM} & .451\std{.00}$^\dagger$ & \textit{n/s} & \textit{n/s} & .432\std{.01} & \textit{n/s} & \textbf{\underline{.529}}\std{.03} & .451\std{.03} & \underline{.517}\std{.00} & \textit{n/s} & .491\std{.01} \\
 & $Q$$\uparrow$ & \underline{.440}\std{.00} & .167\std{.02} & \textit{OOM} & \textit{OOM} & .002\std{.00}$^\dagger$ & \textit{n/s} & \textit{n/s} & \textbf{\underline{.441}}\std{.00} & \textit{n/s} & .014\std{.01} & .298\std{.11} & .354\std{.00} & \textit{n/s} & .053\std{.02} \\
\bottomrule
\end{tabular}}
\end{table*}

\paragraph{\textbf{Large graphs}}
The three largest graphs are where FairPGT is strongest: on Credit, Pokec-n, and Pokec-z, it is the only method with both the highest NMI and the lowest $\Delta_{\mathrm{SP}}$. On Pokec-n and Pokec-z, its $\Delta_{\mathrm{SP}}$ is 0.006 and 0.005, about ten times lower than that of DMoN (0.054 and 0.056), while its NMI is 0.098 and 0.106, against 0.072 and 0.062 for the next best method, FairGT. Since both Pokec graphs have four protected groups, this result also holds beyond binary attributes. On Credit, the NMI gain varies considerably across seeds ($0.042\pm0.053$), whereas the fairness gain is stable ($0.003\pm0.002$).
Modularity shows what this choice means. DMoN finds a highly modular partition of Pokec-n ($Q=0.41$), but its clusters are unrelated to the labels (NMI 0.001) and differ by group ($\Delta_{\mathrm{SP}}=0.054$). FairPGT, in contrast, trades modularity ($Q=0.06$, \autoref{tab:extra}) for clusters that are better aligned with the labels and almost independent of the group, which is exactly the expected outcome from a fair clustering method considering both objectives together.

\paragraph{\textbf{Graphs with group-aligned communities}}
On LastFM and Facebook, the strongest communities are correlated with the protected attribute: DMoN, for example, reaches NMI 0.550 on LastFM, but with $\Delta_{\mathrm{SP}}=0.399$. These are therefore the graphs where fairness costs the most, and FairPGT pays this price openly. It reduces $\Delta_{\mathrm{SP}}$ tenfold on LastFM (0.399 $\to$ 0.040) and more than fourfold on Facebook (0.124 $\to$ 0.027), while keeping NMI at 0.289 and 0.057. Among the methods that reach $\Delta_{\mathrm{SP}}\le0.05$ without collapsing, FairPGT has the highest NMI on both graphs; the only fairer method on LastFM, iFairSC ($\Delta_{\mathrm{SP}}=0.029$), has NMI 0.001. As \autoref{sec:sbm} shows, this loss is the expected price of fairness when communities and groups coincide.

\paragraph{\textbf{Small graphs}}
On the two smallest graphs, however, a classical method remains hard to beat. On NBA, sFairSC is better than FairPGT in both NMI (0.133 vs.\ 0.107) and $\Delta_{\mathrm{SP}}$ (0.011 vs.\ 0.016). On German, no method recovers the labels (NMI $\le0.013$ for all), so German measures fairness only; here too, sFairSC has the lowest $\Delta_{\mathrm{SP}}$ (0.006).

\paragraph{\textbf{Summary}}
Taken together, in mean NMI and $\Delta_{\mathrm{SP}}$, FairPGT is dominated by another method only on NBA, by sFairSC. Most fairness-aware deep baselines (DMoN-SP, FairGT, FairGP) keep most of the unfairness of DMoN on LastFM ($\Delta_{\mathrm{SP}}$ 0.39--0.45), and sFairSC, although it enforces exact parity on the relaxed problem, reaches $\Delta_{\mathrm{SP}}$ 0.37--0.47 on LastFM and Facebook after $k$-means rounding. The fairness of FairPGT, in turn, comes from its assignment-level group term, which acts on the soft assignments throughout training (\autoref{sec:ablation}).

\begin{figure*}[t]
  \centering
  \includegraphics[width=\textwidth]{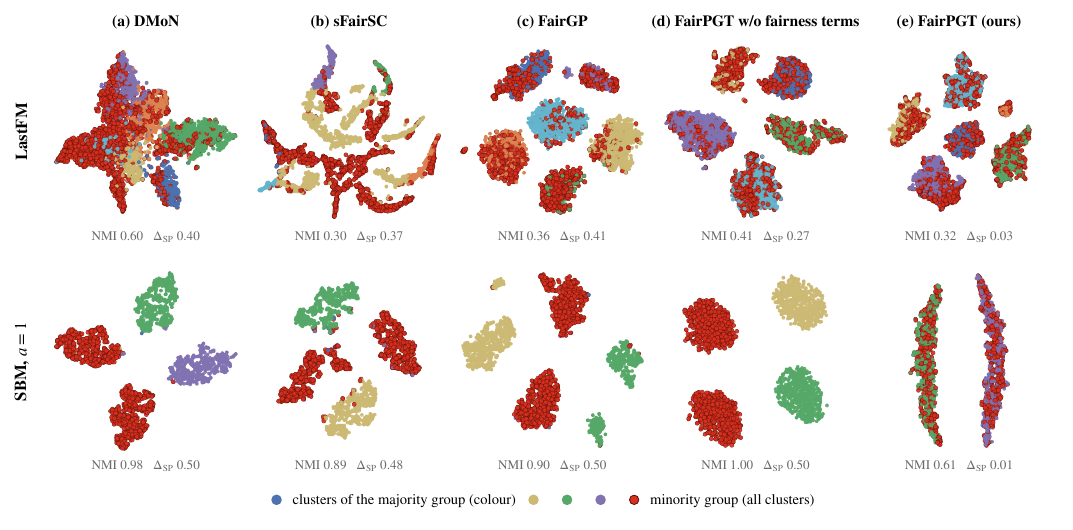}
  \caption{t-SNE of the node embeddings (seed 0). Red: minority group; other colours: the clusters of the majority-group nodes; points are drawn in random order, so a fair clustering shows red spread evenly over all clusters. Methods: the strongest fairness-agnostic baseline (DMoN), the strongest fair baseline on small graphs (sFairSC, its fair spectral embedding), the closest architecture (FairGP), and FairPGT without and with its fairness terms. Below each panel: NMI and $\Delta_{\mathrm{SP}}$ of the plotted run.}
  \label{fig:embeddings}
\end{figure*}

\paragraph{\textbf{Embeddings}}
\autoref{fig:embeddings} shows what this fairness looks like in the learned representations. On the LastFM network, DMoN, sFairSC, and FairGP form clusters in which the minority group is concentrated in some clusters and nearly absent from others, and FairPGT without its fairness terms behaves the same way. With its fairness terms, by contrast, FairPGT keeps well-separated clusters, each of which contains both groups close to their population shares. On the SBM at full alignment, every baseline recovers the four communities, which here are the groups themselves; FairPGT instead forms two clusters that each join one community of each group, which is precisely the structure of the optimal fair clustering in Proposition~\ref{prop:price}. Finally, sFairSC illustrates that a group-neutral representation does not by itself yield fair clusters: its constraint equalizes the group means of the spectral embedding, so a linear probe cannot recover the group from it ($S$-probe 0.50 on LastFM), and yet $k$-means on this embedding yields $\Delta_{\mathrm{SP}}=0.37$.

\subsection{Q2: The price of fairness under controlled alignment}\label{sec:sbm}

Real graphs do not let us control how strongly communities and groups overlap. The SBMs of \autoref{sec:setup}, in contrast, allow these: their alignment $a$ moves from groups that are independent of the communities ($a=0$) to groups that coincide with them ($a=1$), while the features always carry the group. Before looking at the results, it helps to know the best that any fair method can achieve in the extreme case, which the following proposition establishes.

\begin{proposition}[Price of exact parity at full alignment]\label{prop:price}
Let $a=1$, $K=4$ equal communities $Y$, and $S=Y \bmod 2$. Every clustering $C$ with $\Delta_{\mathrm{SP}}=0$ satisfies $\mathrm{NMI}(C,Y)\le 2/3$ (arithmetic normalization), with equality for the pairing $\{0,1\},\{2,3\}$.
\end{proposition}
\begin{proof}
$\Delta_{\mathrm{SP}}=0$ means $C$ is independent of $S$, so $H(S\mid C)=H(S)=1$ bit. Since $S$ is a function of $Y$, $H(Y\mid C)\ge H(S\mid C)=1$, hence $I(C;Y)=H(Y)-H(Y\mid C)\le 2-1=1$. Also $H(C)\ge I(C;Y)$. With $\mathrm{NMI}=2I/(H(Y)+H(C))\le 2I/(2+I)$, which increases in $I$, the bound is $2/3$ at $I=1$. The pairing $\{0,1\},\{2,3\}$ is independent of $S$ and has $I=H(C)=1$.
\end{proof}

\begin{figure*}[t]
  \centering
  \includegraphics[width=\textwidth]{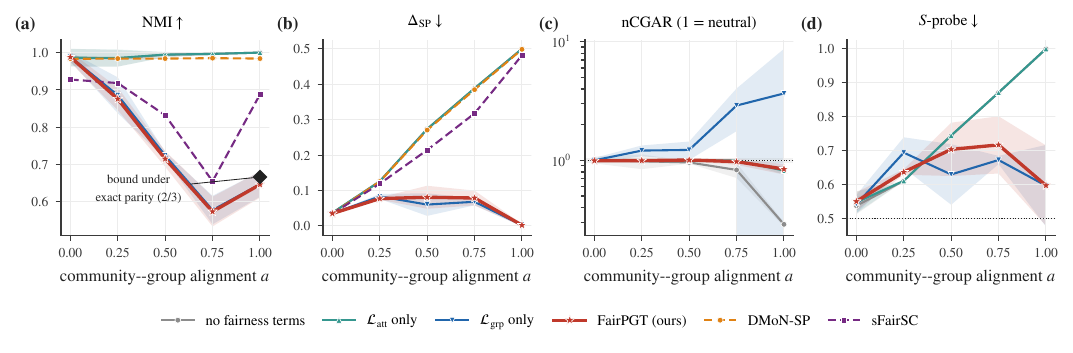}
  \caption{SBMs with community--group alignment $a$ (mean, shaded $\pm$ one std over three seeds). (a)~The diamond marks the NMI bound of \autoref{prop:price} at $a=1$. (c)~nCGAR of the trained encoder (log scale; shaded band: $1\pm0.05$); DMoN-SP and sFairSC have no attention. (d)~$S$-probe: 0.5 means the group cannot be decoded from the embeddings.}
  \label{fig:sbm}
\end{figure*}

\autoref{fig:sbm} shows the results. At $a=0$, all methods recover the communities (NMI $\ge0.92$) and are fair ($\Delta_{\mathrm{SP}}\approx0.035$, the sampling noise of the generator). As $a$ grows, however, the methods without an effective fairness term (FairPGT without fairness terms, FairPGT with the attention term only, and DMoN-SP) keep NMI near 1, while their $\Delta_{\mathrm{SP}}$ rises almost linearly to 0.50; sFairSC follows the same trend (0.48 at $a=1$). FairPGT, by contrast, keeps $\Delta_{\mathrm{SP}}\le0.081$ at every alignment and reaches 0.002 at $a=1$. Its NMI falls to 0.57 at $a=0.75$ and recovers to 0.646 at $a=1$, which is 97\% of the bound in \autoref{prop:price}. Notably, it reaches this value with two clusters of equal size, each joining one community of each group, which is exactly the structure of the optimal fair clustering in the proof of \autoref{prop:price}; on the SBMs, we therefore do not count fewer than $K$ clusters as a collapse. In other words, at full alignment FairPGT is almost as accurate as any exactly fair clustering can be: its loss in NMI is the price of parity itself, not a weakness of the encoder.

\paragraph{\textbf{Encoder neutrality is two-sided}}
Fair clusters, however, do not yet tell us whether the encoder itself is group-neutral. The two right panels of \autoref{fig:sbm} therefore look inside the encoder, and they reveal three distinct behaviours. Without fairness terms, attention follows the groups: at $a=1$, it is strongly group-homophilous (nCGAR 0.29), and the group can be read off the embeddings perfectly ($S$-probe 1.00). With the group term alone, the clusters become fair, but the encoder swings to the other extreme: attention becomes group-heterophilous (nCGAR 2.9 at $a=0.75$ and 3.7 at $a=1$), so the encoder still uses group membership, only in reverse. With both terms, finally, nCGAR stays between 0.84 and 1.01 at every alignment: the clusters are fair, and the attention is neutral. The converse also holds: the nCGAR term alone moves attention toward neutrality (from 0.29 to 0.82 at $a=1$) but leaves $\Delta_{\mathrm{SP}}$ at 0.50, because the features themselves carry the group, by construction of the SBM, and reach the clustering head through the residual path. In short, neither level can stand in for the other.

\begin{figure*}[t]
  \centering
  \includegraphics[width=\textwidth]{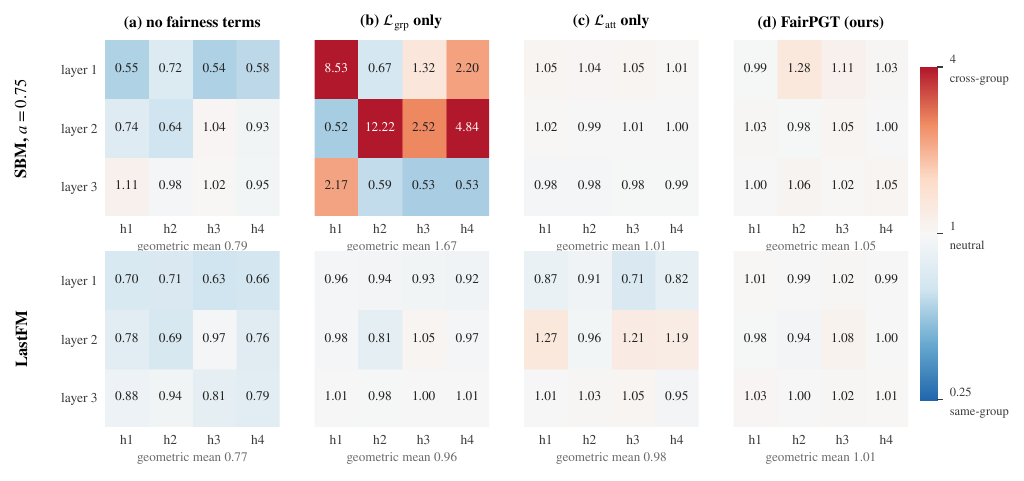}
  \caption{nCGAR of every attention head of the trained encoder (rows: layers; columns: heads), for the METIS partition with the most balanced group composition; 1 is group-neutral, blue cells favour the node's own group, red cells the other group. Top: SBM with alignment $a=0.75$ (the whole partition of 250 nodes). Bottom: LastFM (a stratified sample of 400 of the 682 nodes of the partition). Seed 0.}
  \label{fig:attention}
\end{figure*}

\autoref{fig:attention} takes a closer look, resolving the attention head by head. At $a=0.75$ without fairness terms, every first-layer head is group-homophilous (nCGAR 0.54--0.72). With the group term alone, the heads split: some become strongly heterophilous (up to 12.2), while others remain homophilous (down to 0.52). Its attention thus still depends on group membership, in both directions at once, and the geometric mean of 1.67 hides how far individual heads are from neutral. With the nCGAR term, in contrast, alone or in the full model, all heads stay near 1 (0.98--1.05 and 0.98--1.28, respectively). On LastFM, a real graph, the fairness-agnostic heads are likewise homophilous in the first layer (0.63–0.71), and the full model keeps every head within 0.94–1.08; here, however, the group term alone does not over-correct, as expected from the weaker alignment.

Since nCGAR pools all nodes of a partition, we finally split it by the group of
the attending node. On LastFM, the first layer of the full model is balanced
for both groups: each sends 46--55\% of its attention across groups, against
about 50\% under uniform attention. In the second layer, however, one group
sends 56--62\% and the other 34--49\%, so part of the aggregate balance comes
from the two deviations offsetting each other. On the SBM at $a=0.75$, the
majority group stays close to its uniform share in most heads (15--26\% against
20\%), while the 51 minority nodes of the partition deviate more strongly, in
both directions: in the head with nCGAR 1.28, for example, they send only 2\%
across groups (against 80\% under uniform attention), whereas the majority
sends 48\%. Similarly, we audit the routing branch, which nCGAR does not cover,
by the mutual information between a node's group and the routing token it
attends to. In the full model, this is at most 0.004 nats on every real graph
and 0.046 on the SBM at $a=1$, against 0.62 without fairness terms (the maximum
for two groups is $\log 2 \approx 0.69$). Notably, the attention term alone
leaves it at 0.62, so the routing branch is kept group-independent by the
group term. In short, the trained encoder is balanced in aggregate and its
routing carries almost no group information, but its local attention is not
yet neutral for each group separately.

\subsection{Q3: Ablation}\label{sec:ablation}

\begin{figure*}[t]
  \centering
  \includegraphics[width=\textwidth]{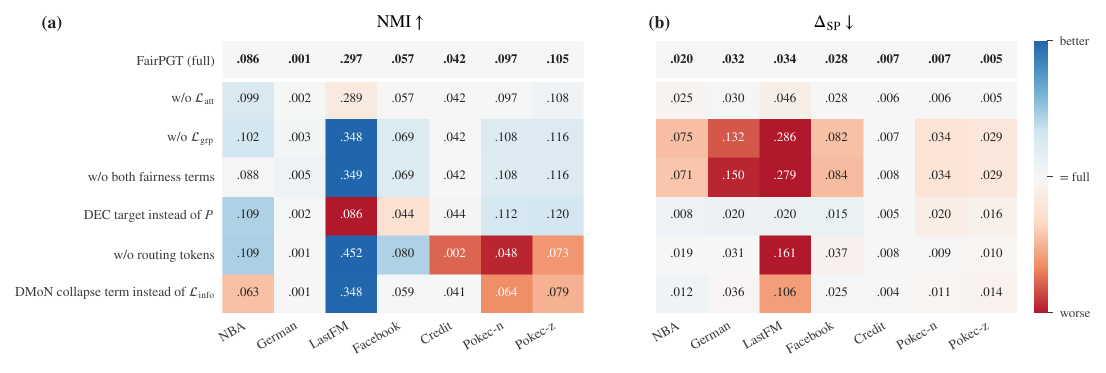}
  \caption{Ablation of FairPGT (mean over five seeds). Each row changes one component of the full model and keeps everything else fixed: w/o $\mathcal{L}_{\mathrm{att}}$ and w/o $\mathcal{L}_{\mathrm{grp}}$ remove one fairness term, w/o both fairness terms removes both; \emph{DEC target instead of $P$} replaces the graph-aware target of Eq.~\eqref{eq:unified-target} by the standard DEC target; \emph{w/o routing tokens} removes cross-partition communication, so attention stays within partitions; \emph{DMoN collapse term instead of $\mathcal{L}_{\mathrm{info}}$} replaces our information term by the collapse regularizer of DMoN. Each cell gives the value of the variant; its colour shows the change relative to the full model (first row), blue where the variant is better and red where it is worse.}
  \label{fig:ablation}
\end{figure*}

To answer Q3, \autoref{fig:ablation} removes or replaces one component at a time, in an independent rerun with the same protocol (differences from \autoref{tab:main} are within one standard deviation). The variants separate the two fairness levels from the two quality terms, and they show that each component has a distinct role.

\begin{itemize}
  \item \emph{The group term ($\mathcal{L}_{\mathrm{grp}}$) is what makes the clusters fair.} Removing it raises $\Delta_{\mathrm{SP}}$ on every dataset except Credit, where it is already 0.007: LastFM 0.034 $\to$ 0.286, German 0.032 $\to$ 0.132, Facebook 0.028 $\to$ 0.082, NBA 0.020 $\to$ 0.075, and Pokec-n 0.007 $\to$ 0.034. Without it, FairPGT is a capable but unfair clustering method.
  \item \emph{The attention term ($\mathcal{L}_{\mathrm{att}}$), in contrast, shapes the encoder, not the outcome.} Removing it changes NMI and $\Delta_{\mathrm{SP}}$ by at most 0.013 on any dataset. On the real graphs, the trained attention is already close to neutral (mean nCGAR 0.90--1.06 in all variants), so the term has little to correct; its effect appears when the alignment is strong (\autoref{sec:sbm}) and at the level of individual heads (\autoref{fig:attention}).
  \item \emph{On the quality side, the graph-aware target ($P$) is what makes FairPGT accurate on strongly modular graphs.} Replacing it with the standard DEC target lowers NMI on LastFM from 0.30 to 0.09 and on Facebook from 0.06 to 0.04, consistent with \autoref{thm:target}: the structural guidance matters where the graph is modular. On the other datasets, NMI changes by at most 0.023. The standard DEC target yields a lower $\Delta_{\mathrm{SP}}$ on five graphs, and on NBA it is better in both metrics; on LastFM and Facebook, however, this fairness comes with the loss of NMI just described, on both Pokec graphs, the standard target gains some NMI (0.112 vs. 0.097 and 0.120 vs. 0.105) but is about three times less fair (0.020 vs. 0.007 and 0.016 vs. 0.005)
  \item \emph{Similarly, routing tokens matter most on large, weakly modular graphs.} Without them, NMI collapses on Credit (0.04 $\to$ 0.00) and drops on Pokec (0.10 $\to$ 0.05--0.07). On the smaller graphs, NMI instead rises (NBA, Facebook, and LastFM, the last to 0.45), but on LastFM only, together with a fivefold larger gap ($\Delta_{\mathrm{SP}}$ 0.034 $\to$ 0.161). In other words, letting partitions communicate with each other is what keeps FairPGT accurate on the large graphs.
    \item \emph{Finally, our information term matters for fairness, not only for quality.} A common alternative for preventing degenerate clusters is the collapse regularizer of DMoN~\citep{DBLP:journals/jmlr/TsitsulinPPM23}, $\sqrt K\|\bar c\|_2-1$, which constrains only the cluster sizes and, unlike $\mathcal{L}_{\mathrm{info}}$, is already minimal when all nodes share the uniform assignment. To test whether our choice matters, we replace $\mathcal{L}_{\mathrm{info}}$ with this regularizer and keep everything else fixed. The replacement lowers NMI on NBA (0.09 $\to$ 0.06) and Pokec (0.10 $\to$ 0.06--0.08), and triples $\Delta_{\mathrm{SP}}$ on LastFM (0.034 $\to$ 0.106); its higher NMI on LastFM (0.35) comes only together with that larger gap.
\end{itemize}

Taken together, the ablation mirrors the design of FairPGT: two terms make it fair, at the level of the clusters and of the attention, and two components make it accurate, the graph-aware target on modular graphs and the routing tokens on large ones. None of them can be dropped without losing something the others do not recover.

\subsection{Q4: Sensitivity and cost}\label{sec:cost}

Finally, we turn to the practical side: how sensitive FairPGT is to its weights, and what partitioning costs and buys.

\paragraph{\textbf{Fairness weight}}
The weight $\lambda_{\mathrm{grp}}$ is the main control of the trade-off (three seeds per value; red line in \autoref{fig:tradeoff}). On LastFM, small weights ($\lambda_{\mathrm{grp}}\le0.3$) leave the gap almost unchanged, and from $\lambda_{\mathrm{grp}}=1$ upward $\Delta_{\mathrm{SP}}$ falls steadily, from 0.260 at 1 to 0.125 at 3 and 0.035 at 10, while NMI decreases from 0.362 at $\lambda_{\mathrm{grp}}=0$ to 0.288. On NBA, by contrast, fairness costs no measurable accuracy: $\Delta_{\mathrm{SP}}$ falls from 0.103 to 0.017 while NMI stays within seed variation (0.056–0.093 across all weights). The attention weight, in turn, is robust: $\lambda_{\mathrm{att}}\in[0,10]$ changes neither NMI nor $\Delta_{\mathrm{SP}}$ beyond seed variation (LastFM: NMI 0.285--0.301, $\Delta_{\mathrm{SP}}$ 0.032--0.048), in line with its role of shaping the encoder rather than the outcome. Finally, $\lambda_{\mathrm{info}}$ is best kept moderate: without it ($\lambda_{\mathrm{info}}=0$), the gap on LastFM more than doubles (0.042 $\to$ 0.098), while at $\lambda_{\mathrm{info}}=3$ NMI on LastFM drops to 0.228, so values in $[0.3,1]$ balance the two.

\paragraph{\textbf{Partitions}}
\autoref{fig:scaling} varies the number of partitions $N_p$ (one seed, 50 epochs). As predicted by \autoref{thm:complexity}, peak memory follows $1/N_p$ closely at small $N_p$ and flattens once the per-node terms dominate: it falls from 11.9\,GB to 0.75\,GB on LastFM ($N_p=1\to64$), from 50\,GB to 5.5\,GB on Facebook, and from 88\,GB to 8.1\,GB on Credit ($N_p=4\to64$), while Pokec does not fit at $N_p \leq 16$ and fits from $N_p = 32$ (our default $N_p = 27$ needs 72 GB, Table 5). Time per epoch falls accordingly, by up to $7\times$. Importantly, restricting attention to partitions does not cost quality: on LastFM, NMI even rises with $N_p$, from 0.16 at $N_p=1$ to 0.34--0.38 at $N_p\ge32$, and on Facebook, Credit, and Pokec it changes by at most 0.04.

\paragraph{\textbf{Cost}}
FairPGT is heavier than lightweight GNN clustering methods: with the default $N_p$, a run on Pokec takes 94\,s and 72\,GB, against 1\,s and 0.7\,GB for DMoN and 9\,s and 3.6\,GB for FairGP (\autoref{tab:cost}). It is, however, faster than several deep clustering baselines on the same GPU, taking 38\,s on Facebook against 71\,s for MinCutPool and 752\,s for FairGC, and it completes on Pokec, where CCGC and SCGC run out of memory. Moreover, memory, its main cost, is directly controlled by $N_p$: on Pokec, 64 partitions need less than half the memory of the default (\autoref{fig:scaling}) wit similar NMI.

\begin{table*}[t]
\centering
\scriptsize
\caption{Training time in seconds$\downarrow$ / peak GPU memory in GB$\downarrow$ (200 epochs, mean over five seeds; memory is the peak allocated by PyTorch). --: skipped or out of memory. Runs used different GPUs (\autoref{sec:setup}), so times are indicative.}
\label{tab:cost}
\adjustbox{max width=\textwidth}{
\begin{tabular}{lccccccc}
\toprule
Method & NBA & German & LastFM & Facebook & Credit & Pokec-n & Pokec-z \\
\midrule
DMoN & 0.7 / 0.03 & 0.7 / 0.03 & 0.7 / 0.26 & 2.1 / 4.51 & 4.6 / 7.24 & 1.0 / 0.74 & 1.2 / 0.86 \\
SDCN & 3.2 / 0.41 & 3.1 / 0.34 & 6.9 / 1.04 & 74.0 / 18.61 & 16.1 / 9.48 & 52.8 / 19.83 & 60.4 / 23.14 \\
CCGC & 1.0 / 0.07 & 0.9 / 0.06 & 1.6 / 0.87 & 19.4 / 14.67 & 10.4 / 25.36 & -- & -- \\
SCGC & 6.6 / 0.03 & 15.1 / 0.06 & 85.8 / 1.04 & 601.4 / 17.32 & 228.8 / 29.18 & -- & -- \\
DinkNet & 1.3 / 0.07 & 1.2 / 0.06 & 2.1 / 0.42 & 8.7 / 6.74 & 12.7 / 11.08 & 56.8 / 36.03 & 59.3 / 37.37 \\
DiffPool & 0.7 / 0.02 & 0.6 / 0.05 & 1.7 / 0.98 & 10.3 / 16.66 & 17.2 / 28.89 & -- & -- \\
MinCutPool & 1.1 / 0.03 & 1.2 / 0.04 & 4.2 / 0.51 & 71.3 / 8.61 & 94.2 / 14.51 & -- & -- \\
\midrule
DMoN-SP & 1.0 / 0.03 & 1.0 / 0.03 & 1.5 / 0.26 & 1.8 / 4.55 & 2.9 / 7.24 & 2.0 / 0.74 & 1.3 / 0.86 \\
FairGC & 1.4 / 0.04 & 1.3 / 0.10 & 25.6 / 2.04 & 752.5 / 33.25 & -- & -- & -- \\
FairGT & 1.9 / 0.07 & 1.6 / 0.06 & 2.5 / 0.26 & 4.7 / 4.51 & 3.6 / 7.24 & 8.1 / 2.67 & 9.3 / 3.17 \\
FairGP & 1.7 / 0.03 & 1.7 / 0.03 & 2.4 / 0.26 & 3.9 / 4.55 & 8.8 / 7.24 & 8.9 / 3.63 & 9.7 / 4.21 \\
sFairSC & 0.0 / 0.00 & 0.2 / 0.00 & 0.2 / 0.12 & 3.3 / 2.44 & 6.7 / 3.60 & 5.4 / 0.03 & 5.1 / 0.04 \\
iFairSC & 0.2 / 0.02 & 0.7 / 0.02 & 56.2 / 0.14 & -- & -- & -- & -- \\
\midrule
\textbf{FairPGT} & 5.8 / 0.12 & 4.5 / 0.20 & 9.5 / 1.92 & 37.8 / 24.92 & 43.3 / 32.69 & 94.1 / 71.56 & 93.3 / 71.72 \\
\bottomrule
\end{tabular}}
\end{table*}

In summary, FairPGT has a single weight that steers the trade-off, two weights that need no tuning within a broad range, and one structural knob, $N_p$, that reduces memory with little change in NMI over the tested settings.

\subsection{Discussion, limitations, and outlook}\label{sec:discussion}

\paragraph{\textbf{Fairness in assignments and attention}}
Taken together, the experiments separate two levels of fairness. Outcome fairness comes from the assignment-level group term: it lowers the statistical-parity gap wherever there is a gap to lower, that is, on every dataset except Credit, where the gap is already 0.007. Attention-level balance, in turn, comes from the nCGAR term, which discourages both same-group and cross-group preference in the aggregate partition-local attention but does not make the clusters fair on its own. The reason is architectural: a residual connection carries each node's own features, which can encode the group, past any attention pattern. Aggregate attention balance is therefore a property of the encoder that can be audited separately from the fairness of the assignments, rather than a guarantee on the outcome. This is exactly the role in which we use nCGAR: as a diagnostic of the trained encoder, and as a regularizer that counters the attention imbalance the group term produces when used alone. This separation is, in our view, the main lesson of this work.

\begin{figure*}[t]
  \centering
  \includegraphics[width=\textwidth]{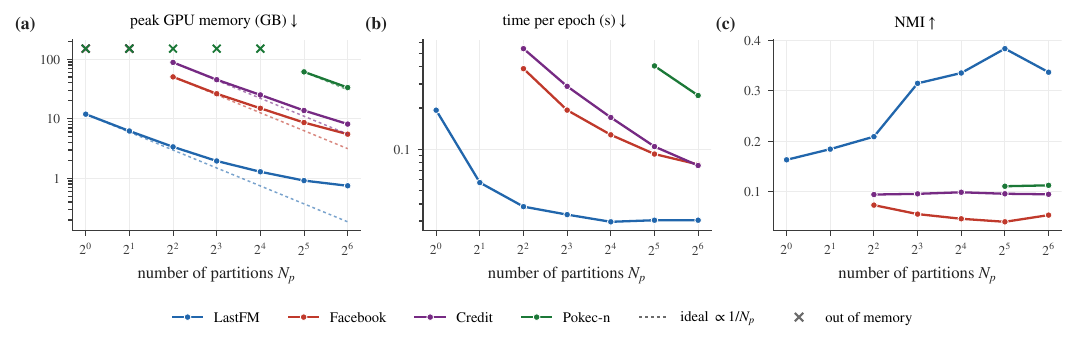}
  \caption{Effect of the number of METIS partitions $N_p$ (one seed, 50 epochs). Dashed lines in (a) show the ideal $1/N_p$ scaling from each dataset's first feasible point; crosses mark runs that exceed the GPU memory (Facebook and Credit at $N_p \leq 2$, Pokec at $N_p \leq 16$).}
  \label{fig:scaling}
\end{figure*}

\paragraph{\textbf{Limitations and future work}}
Like any method, FairPGT has a scope that our experiments delineate, and each part of it points to a concrete next step. First, FairPGT does not maximize modularity: on Pokec, its clusters are less modular than those of DMoN ($Q=0.06$ vs.\ $0.41$) but follow the labels more closely, so when the most modular partition is itself the goal, a modularity-based method remains the better choice. Second, on the two smallest graphs, fair spectral clustering remains a strong competitor, which suggests that the benefits of attention grow with the size and richness of the graph. Third, attention within partitions makes FairPGT more demanding than lightweight GNN methods; the number of partitions, however, offers a direct way to reduce its memory with little change in NMI over the tested settings (\autoref{fig:scaling}). Fourth, nCGAR balances the aggregate attention of each partition, not separately for each group; the per-group diagnostic of \autoref{sec:sbm} makes this visible, so practitioners can check it on any trained encoder. Fifth, FairPGT is transductive: partitions and spectral encodings are computed on the whole graph, so adding new nodes currently requires rerunning the pipeline. Finally, the method handles any number of categorical groups, so several attributes can be combined into intersectional groups (for example, gender and four age groups give $G=8$). Our experiments use one protected attribute at a time, binary on five graphs and with four groups on Pokec; evaluating intersectional settings, in which small intersections can make the group shares noisy, is a natural next step.

\paragraph{\textbf{Ethical considerations}}
FairPGT is meant for settings where clusters inform decisions about people, such as customer segmentation in finance or community analysis in social networks, and three cautions apply. First, it needs the protected attribute of every training node. Such attributes are personal data, some of them legally protected, so collecting and storing them requires a legal basis and safeguards; the trained model does not use them at inference, which confines their use to training. Second, statistical parity is one notion of fairness among several: clusters with equal group shares can still feed unfair decisions downstream, parity on one attribute does not imply parity for intersections of attributes, and fair clusters must not be used to justify treating groups differently. Third, the groups of LastFM and Facebook are constructed from the features and serve only as benchmarks, without demographic meaning. Conversely, the nCGAR of a trained encoder offers practitioners a way to audit whether a graph transformer's attention mixes the groups evenly, independently of how its output is later used.

\section{Conclusion}\label{sec:conclusion}

We presented FairPGT, a partitioned graph transformer for fair unsupervised graph clustering. Partition-local attention with routing tokens makes a transformer encoder feasible on graphs with tens of thousands of nodes, and a graph-aware self-training target lets it exploit the community structure. On top of this, FairPGT imposes fairness at two complementary levels. The assignment-level term makes the clusters fair, and on the three largest benchmarks FairPGT has both the highest mean NMI and the lowest statistical-parity gap. The nCGAR term, in turn, keeps the encoder's attention balanced across groups; when communities and groups align strongly, the assignment term alone drives the attention to become group-heterophilous. Moreover, on controlled graphs where communities and groups coincide, FairPGT reaches 97\% of the best NMI that any exactly fair clustering can achieve, so its loss in quality there is largely the price of fairness itself rather than a weakness of the encoder. Above all, our results show that fair clusters and balanced attention are distinct goals, and that a fair clustering method should, and can, reach both. Two extensions follow naturally from this work: an inductive version for evolving graphs and unseen nodes, and control of the quality--fairness trade-off at inference time.

\section*{Acknowledgments}
Siamak Ghodsi gratefully acknowledges support from the German Research Foundation (DFG SPP 2443 (543081196)).
Amjad Seyedi acknowledges support from the European Union (ERC consolidator, eLinoR, no. 101085607).

\bibliography{References}

\newpage

\end{document}